\documentclass{article}

\usepackage[main, final]{neurips_2026}

\usepackage[utf8]{inputenc}
\usepackage[T1]{fontenc}
\usepackage{hyperref}
\usepackage{url}
\usepackage{booktabs}
\usepackage{amsfonts}
\usepackage{amsmath}
\usepackage{amssymb}
\usepackage{nicefrac}
\usepackage{microtype}
\usepackage{xcolor}
\usepackage{graphicx}
\usepackage{algorithm}
\usepackage{algpseudocode}
\usepackage{enumitem}
\usepackage{multirow}
\usepackage{subcaption}
\usepackage{mathtools}
\usepackage{xspace}
\usepackage{amsthm}
\usepackage{mathtools}
\usepackage{microtype}
\usepackage{xcolor}
\usepackage{graphicx}
\usepackage{float}
\usepackage{caption}
\usepackage{subcaption}
\usepackage{wrapfig}
\usepackage{colortbl}
\usepackage{pifont}
\usepackage{wrapfig}
\usepackage{caption}
\usepackage{algorithm}
\usepackage{algpseudocode}
\usepackage{xcolor}
\usepackage{tabularx}
\usepackage{tcolorbox}

\newcommand{\bx}{\boldsymbol{x}}

\newcommand{\calD}{\mathcal{D}}

\newcommand{\calE}{\mathcal{E}}
\newcommand{\calH}{\mathcal{H}}
\newcommand{\calS}{\mathcal{S}}
\newcommand{\E}{\mathbb{E}}

\newcommand{\by}{\mathbf{y}}
\newcommand*{\modelname}{\text{LLM-AutoSciLab}\@\xspace}
\newcommand*{\benchmark}{\text{ActiveSciBench}\@\xspace}
\newcommand*{\chem}{\text{ActiveSciBench-Chem}\@\xspace}
\newcommand*{\grn}{\text{ActiveSciBench-GRN}\@\xspace}

\newcommand{\cmark}{\ding{51}}  
\newcommand{\xmark}{\ding{55}}  
\definecolor{darkgreen}{rgb}{0.0, 0.7, 0.0}

\title{\modelname: Closed-Loop Scientific Discovery via Active Experimentation with LLMs}
\vspace{-0.2in}
\author{%
}

\author{\fontsize{9.750}{10}\selectfont
Sanchit Kabra$^{1}$\textsuperscript{*}, 
Nikhil Abhyankar$^{1}$\textsuperscript{*}, 
Saaketh Desai$^{2}$, 
Prasad P. Iyer$^{2}$, 
Chandan K.\ Reddy$^{1}$ \\[0.5em]
{\small $^{1}$Virginia Tech} ~~~~~~~ \small $^{2}$Sandia National Laboratories}

\begin{document}

\maketitle
\begingroup
\renewcommand\thefootnote{}
\footnotetext{\hspace{-0.45em}\textsuperscript{*}Equal contribution. Correspondence: \texttt{sanchit23@vt.edu, nikhilsa@vt.edu}.}
\endgroup
\vspace{-0.2in}
\begin{abstract}
Scientific discovery is a closed-loop process where hypotheses guide data acquisition and observations refine the hypothesis space. Yet most approaches reduce discovery to supervised learning over fixed datasets, where limited observations can support multiple plausible mechanisms that fit locally but fail to generalize. Thus, the key challenge is selecting informative observations to resolve uncertainty, shifting the focus from static inference to adaptive data acquisition. To address this, we propose \textbf{\modelname}, a closed-loop framework that couples hypothesis generation with hypothesis-conditioned experiment selection and mechanism refinement. Rather than fitting models to passively collected data, \modelname iteratively proposes plausible hypotheses, selects informative experiments to distinguish or refine them, and updates its state using the resulting evidence. \textcolor{black}{To evaluate dynamic, closed-loop scientific discovery with active data acquisition, we introduce \textbf{\benchmark}, comprising two datasets: (i) {\chem} (57 enzyme-kinetics tasks) and (ii) {\grn} (45 gene-regulatory-network tasks),  that model discovery as a budget-constrained process requiring adaptive experiment design, variable selection, and recovery of true mechanisms.} Across NewtonBench, \chem, and \grn, \modelname outperforms prior methods, achieving 67.6\% and 35.1\% symbolic accuracy on NewtonBench and \chem, respectively, and 31.1\% exact graph recovery on \grn. Moreover, hypothesis-guided experimentation is 2--5$\times$ more sample-efficient than the strongest competing baselines.\footnote{Code: https://github.com/scientific-discovery/LLM-AutoSciLab}

\end{abstract}

\vspace{-1.em}
\section{Introduction}
\label{sec:intro}
\vspace{-0.5em}
Discovering governing principles underlying physical systems remains a central challenge in science~\citep{udrescu2020ai,petersen2021deep}. Recent advances in large language models (LLMs) have enabled systems that leverage pretrained knowledge, reasoning, and tool use to generate hypotheses, analyze observations, and accelerate scientific discovery~\citep{wang2023scientific,ai4science2023impact,reddy2025towards}. However, \emph{existing methods treat discovery as static, supervised inference on fixed datasets}~\citep{cranmer2023interpretable, shojaee2025llmsr}. This static formulation creates an identifiability bottleneck, where multiple competing hypotheses can fit the limited observed data equally well, while failing to generalize, making it impossible to recover the true underlying law~\citep{jiang2025active}.

In practice, scientific discovery is inherently a closed loop, with hypotheses guiding experiments and observations refining subsequent hypotheses~\citep{chen2025ai4research}. Crucially, scientists design experiments to induce targeted variations that force competing explanations to diverge, revealing distinctions that static data cannot resolve~\citep{box1967discrimination, ouyang2016practical}. Although self-driving laboratories (SDLs) and active learning systems enable adaptive experimentation~\citep{ling2017high, kusne2020fly, desai2025autoscilab}, they still require substantial human effort for hypothesis formulation and refinement. Moreover, their acquisition strategies are typically optimized for predictive performance and uncertainty reduction, rather than mechanism identification. Consequently, they are not designed to actively resolve competing hypotheses, limiting recovery of the true underlying law under constrained experimental budgets.


\begin{figure}[!htbp]
    \vspace{-1em}
    \centering
    \includegraphics[width=\linewidth]{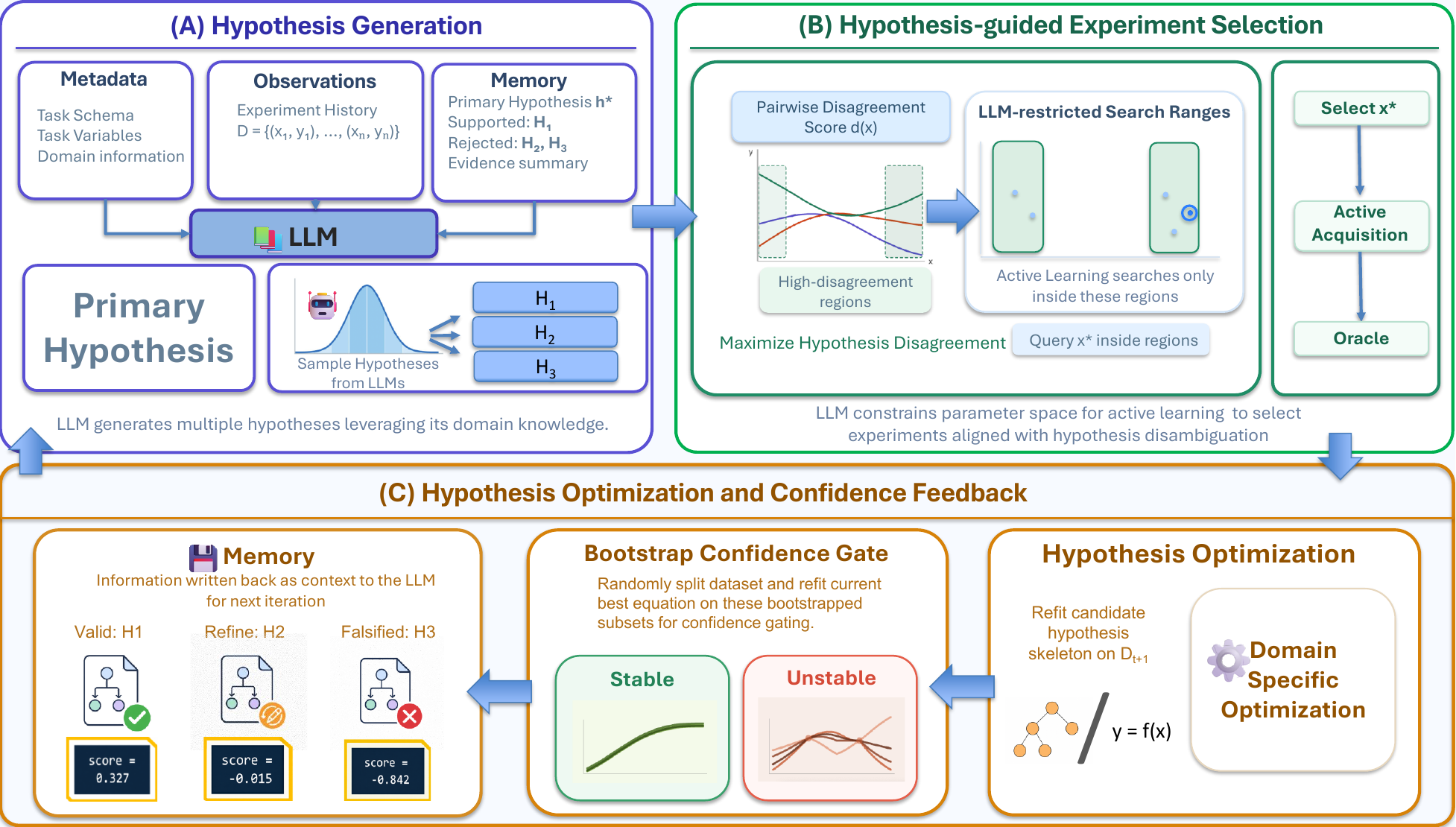}
    \vspace{-1em}
    \caption{\small \textbf{Overview of \modelname.} (A) An LLM generates candidate hypotheses from observations and memory. (B) Experiments are actively selected in regions of maximal disagreement with the hypothesis. (C) Candidates are iteratively refined via domain-specific optimization (e.g., parameter fitting and constraint enforcement), with confidence-based feedback guiding updates.}
    \label{fig:main_fig}
    \vspace{-1em}
\end{figure}
\vspace{-1pt}
To address this gap, we propose \textbf{\modelname}, \emph{a closed-loop framework that models scientific discovery as active hypothesis-conditioned experiment design rather than passive regression over fixed datasets}~(Table~\ref{tab:comparison}). At iteration $t$, \modelname constructs a structured mechanism hypothesis set from accumulated observations and previous interactions, then identifies regions where candidate mechanisms are predicted to disagree. New experiments are selected online using a \emph{hypothesis-conditioned acquisition objective that prioritizes mechanism disambiguation}, acquiring data most informative for separating competing laws~(Figure~\ref{fig:main_fig}). The resulting observation is used to evaluate, refine, or eliminate hypotheses, and update the next acquisition step. Unlike Bayesian or traditional active learning methods that acquire data to reduce uncertainty, \modelname selects experiments to maximize disagreement among explicit candidate mechanisms, enabling law recovery under constrained experimental budgets.

Real-world closed-loop discovery requires evaluation settings in which data is actively acquired through experimental design. As shown in Table~\ref{tab:benchmark_comparison}, existing benchmarks~\citep{udrescu2020ai, cranmer2023interpretable, shojaee2025llmsrbench} assume fully observed, fixed datasets, reducing discovery to static function fitting. NewtonBench~\citep{zheng2026newtonbench} introduces interactive probing of memorization-resistant counterfactual laws, but remains limited to predefined input-output physics variables and symbolic law recovery. \textcolor{black}{We address this gap by introducing \textbf{\benchmark}, a benchmark suite for active experimental design grounded across two scientific domains: chemistry and gene regulatory networks.}
Both datasets impose budget-limited oracle access, in which relevant variables are hidden and must be discovered jointly with the experimental design and hypothesis refinement. \chem focuses on enzyme-kinetic rate laws from selected reaction conditions with distractor variables, while \grn targets signed causal regulatory graphs from perturbation-response experiments. Together, they move evaluation beyond symbolic regression to both equation-structured and graph-structured discovery. We evaluate \modelname using \texttt{GPT-4o-mini} and \texttt{Qwen-3-4B/14B/32B}, demonstrating that it discovers governing mechanisms faster across settings. Our main contributions can be summarized as:\vspace{-0.2em}
\begin{itemize}[leftmargin=*]
\setlength\itemsep{-0.2em}
    \item We introduce \modelname, a \textit{closed-loop scientific discovery framework} coupling LLM-guided hypothesis generation, hypothesis-conditioned experiment design, and refinement.
    \textcolor{black}{\item We introduce {\benchmark}, \textit{a benchmark suite for active sequential discovery in scientifically grounded systems}, where data is acquired under budget-limited oracle access and relevant variables must be identified.}
    \item We propose a \textit{hypothesis-conditioned acquisition strategy} that selects experiments maximizing disagreement among competing hypotheses, improving sample efficiency under fixed budgets.
    \item We show that \modelname outperforms prior methods across benchmarks, achieving up to 67.6\% symbolic accuracy and 31.1\% exact graph recovery, improving sample efficiency by 2$-$5$\times$. Ablations confirm the importance of each component.
\end{itemize}
\vspace{-1.em}
\begin{table}[!htbp]
\centering
\vspace{-.5em}
\caption{\small \textcolor{black}{Comparison of scientific discovery frameworks across key design dimensions.}}
\vskip 0.01in
\resizebox{\textwidth}{!}{
\begin{tabular}{lccccc}
\toprule
\multirow{2}{*}{\textbf{Method / Framework}} 
& \textbf{Experimental} 
& \textbf{Feedback-driven} 
& \textbf{Closed-Loop} 
& \textbf{Representation} 
& \textbf{Primary} \\
& \textbf{Design}
& \textbf{Refinement} 
& \textbf{Discovery} 
& \textbf{Agnostic}  
& \textbf{Output} \\
\midrule

Passive law discovery~\citep{udrescu2020ai, cranmer2023interpretable}
& \textcolor{red}{\xmark} 
& \textcolor{red}{\xmark} 
& \textcolor{red}{\xmark} 
& \textcolor{red}{\xmark} 
& Equation \\

LLM-guided equation discovery~\citep{shojaee2025llmsr, behzadifar2025decompose}
& \textcolor{red}{\xmark} 
& \textcolor{darkgreen}{\cmark} 
& \textcolor{red}{\xmark} 
& \textcolor{red}{\xmark} 
& Equation  \\

General experiment design~\citep{ouyang2016practical, kusne2020fly}
& \textcolor{darkgreen}{\cmark}  
& \textcolor{darkgreen}{\cmark}
& \textcolor{red}{\xmark} 
& \textcolor{red}{\xmark} 
& Query policy  \\

Domain-specific adaptive discovery~\citep{ling2017high, Bailey_2024}
& \textcolor{darkgreen}{\cmark}  
& \textcolor{darkgreen}{\cmark}
& \textcolor{red}{\xmark} 
& \textcolor{red}{\xmark} 
& Optimized condition \\

Self-driving laboratories~\citep{Abolhasani2023TheRO, doi:10.1126/sciadv.aaz8867}
& \textcolor{darkgreen}{\cmark} 
& \textcolor{darkgreen}{\cmark} 
& \textcolor{darkgreen}{\cmark} 
& \textcolor{red}{\xmark} 
& Domain objective \\

Interpretable SDLs~\citep{desai2025autoscilab, Langley_2024}
& \textcolor{darkgreen}{\cmark} 
& \textcolor{darkgreen}{\cmark}
& \textcolor{darkgreen}{\cmark} 
& \textcolor{red}{\xmark} 
& Equation~/~model \\

Biological network inference~\citep{Pratapa642926, 10.1093/bioinformatics/btr373}
& \textcolor{red}{\xmark} 
& \textcolor{darkgreen}{\cmark}
& \textcolor{red}{\xmark} 
& \textcolor{red}{\xmark} 
& Causal graph \\

\midrule
\cellcolor{yellow!15}\bf {\modelname} (Ours) 
& \cellcolor{yellow!15}\textcolor{darkgreen}{\cmark}
& \cellcolor{yellow!15}\textcolor{darkgreen}{\cmark}
& \cellcolor{yellow!15}\textcolor{darkgreen}{\cmark}
& \cellcolor{yellow!15}\textcolor{darkgreen}{\cmark}
& \cellcolor{yellow!15}\textbf{Open-ended mechanism} \\

\bottomrule
\end{tabular}
}
\vspace{-1em}
\label{tab:comparison}
\end{table}

\vspace{-0.5em}
\section{Related Work}
\vspace{-0.5em}
\label{sec:related}

\paragraph{LLMs for Scientific Discovery.}
LLMs have shown strong potential for accelerating scientific discovery through embedded knowledge and reasoning for hypothesis generation~\citep{zhou2024hypothesis, jansen2026generating}, data-driven analysis~\citep{majumder2024data, reddy2025towards,  agarwal2025autodiscovery}, and equation discovery~\citep{shojaee2025llmsr, grayeli2024symbolic, behzadifar2025decompose}. LLM-based discovery frameworks have also been applied across domains such as chemistry~\citep{wang2025efficient}, materials discovery~\citep{abhyankar2026llema, gan2025matllmsearch}, and program synthesis~\citep{romera2024mathematical}. However, most existing systems are both representation-specific and passive, searching within a predetermined output space, such as equations, materials, or programs, and use LLMs primarily for post-hoc hypothesis generation and refinement over pre-collected, static datasets~(Table~\ref{tab:comparison}). We extend this line of work by using the representational flexibility of LLMs beyond candidate generation, where hypotheses serve as mechanism-level objects that guide online experiment selection, closing the loop between hypothesis generation, data acquisition, and refinement.


\vspace{-.5em}
\paragraph{Experiment Design for Scientific Discovery.}
Experiment design formalizes discovery as selecting measurements that reduce uncertainty over hypotheses under limited budgets~\citep{ouyang2016practical}, with applications across materials and process optimization~\citep{ling2017high, kusne2020fly}, drug discovery and molecular design~\citep{Bailey_2024, Kyro_2024}, genomics and perturbation screening~\citep{Huang2023.12.12.571389, qin2024active}, and applied physics~\citep{doi:10.1073/pnas.1714936115}. Self-driving laboratories extend this paradigm to physical closed-loop platforms by coupling adaptive decision-making with automated synthesis and characterization~\citep{Abolhasani2023TheRO, doi:10.1126/sciadv.aaz8867}. Recent systems such as AutoSciLab~\citep{desai2025autoscilab} integrate active learning with symbolic model recovery, while broader discovery frameworks coordinate experiment selection, model construction, and revision~\citep{Langley_2024}. However, such systems often depend on domain-specific experimental interfaces, acquisition objectives, model classes, or predefined hypothesis spaces, limiting representation-agnostic discovery. \modelname instead treats acquisition as mechanism discrimination: it constructs competing hypotheses, identifies where they diverge, and selects experiments to separate, refine, or falsify them.
\vspace{-.5em}

\paragraph{Benchmarks for Scientific Discovery.}
Scientific discovery benchmarks have largely evaluated recovery from fixed observations, where variables are provided, and the target is an equation or predictive model~\citep{udrescu2020ai, cranmer2023interpretable}~(Table~\ref{tab:benchmark_comparison}). Recent discovery benchmarks reduce memorization through newly generated or out-of-distribution tasks, but still evaluate discovery as offline recovery from pre-collected datasets rather than active acquisition of informative observations~\citep{shojaee2025llmsrbench, kabra2026surfacebench}. NewtonBench~\citep{zheng2026newtonbench} introduces active querying over counterfactual systems, but remains limited to predefined variables and closed-form law recovery. Other benchmarks focus on dynamical prediction~\citep{takamoto2022pdebench, d'ascoli2024odeformer}, condition optimization~\citep{hase2021olympus}, or causal and gene-regulatory inference from benchmark-provided perturbations~\citep{chevalley2025large, Pratapa642926, 10.1093/bioinformatics/btr373}. In contrast, our benchmarks evaluate active mechanism discovery, where the learner must choose experiments under a fixed budget, identify relevant variables, and recover equation- or graph-structured mechanisms from hidden experimental systems.
\begin{table}[!htbp]
\centering
\vspace{-1em}
\caption{\small Comparison of scientific discovery benchmarks across key properties.}
\vspace{0.1em}
\resizebox{\textwidth}{!}{
\begin{tabular}{lcccccc}
\toprule

\multirow{2}{*}{\textbf{Benchmark / Suite}} 
& \textbf{Active} 
& \textbf{Perturbational /} 
& \textbf{Physical} 
& \textbf{Variable}
& \textbf{Memorization}
& \textbf{Discovery} \\

& \textbf{Exploration} 
& \textbf{Interventional} 
& \textbf{Grounding} 
& \textbf{Selection}
& \textbf{Free}
& \textbf{Target} \\

\midrule

Passive law benchmarks~\citep{udrescu2020ai, cranmer2023interpretable}
& \textcolor{red}{\xmark} 
& \textcolor{red}{\xmark} 
& \textcolor{darkgreen}{\cmark} 
& \textcolor{red}{\xmark} 
& \textcolor{red}{\xmark}
& Closed-form law \\

LLM equation benchmarks~\citep{shojaee2025llmsrbench}
& \textcolor{red}{\xmark} 
& \textcolor{red}{\xmark}
& \textcolor{red}{\xmark}
& \textcolor{red}{\xmark} 
& \textcolor{darkgreen}{\cmark}
& Out-of-Distribution \\

Interactive law discovery~\citep{zheng2026newtonbench}
& \textcolor{darkgreen}{\cmark}
& \textcolor{red}{\xmark}
& \textcolor{red}{\xmark}
& \textcolor{red}{\xmark} 
& \textcolor{darkgreen}{\cmark}
& Counterfactual law \\

Dynamical-system benchmarks~\citep{takamoto2022pdebench, d'ascoli2024odeformer}
& \textcolor{red}{\xmark}
& \textcolor{red}{\xmark}
& \textcolor{darkgreen}{\cmark}
& \textcolor{red}{\xmark}
& \textcolor{darkgreen}{\cmark}
& Dynamical behavior \\

Optimization benchmarks~\citep{hase2021olympus}
& \textcolor{darkgreen}{\cmark}
& \textcolor{darkgreen}{\cmark}
& \textcolor{darkgreen}{\cmark}
& \textcolor{red}{\xmark}
& \textcolor{darkgreen}{\cmark}
& Condition optimization \\

Causal perturbation benchmarks~\citep{chevalley2025large}
& \textcolor{red}{\xmark}
& \textcolor{darkgreen}{\cmark}
& \textcolor{darkgreen}{\cmark}
& \textcolor{red}{\xmark}
& \textcolor{darkgreen}{\cmark}
& Causal structure \\

GRN inference benchmarks~\citep{Pratapa642926, 10.1093/bioinformatics/btr373}
& \textcolor{red}{\xmark}
& \textcolor{darkgreen}{\cmark}
& \textcolor{darkgreen}{\cmark}
& \textcolor{red}{\xmark}
& \textcolor{darkgreen}{\cmark}
& GRN inference \\

\midrule

\rowcolor{yellow!15}
\textbf{\benchmark (Ours)}
& \textcolor{darkgreen}{\cmark}
& \textcolor{darkgreen}{\cmark}
& \textcolor{darkgreen}{\cmark}
& \textcolor{darkgreen}{\cmark}
& \textcolor{darkgreen}{\cmark}
& \textbf{Active mechanism discovery} \\

\bottomrule
\end{tabular}}
\vspace{-1em}
\label{tab:benchmark_comparison}
\end{table}

\section{\textcolor{black}{\modelname Method}}
\label{sec:method}
\vspace{-0.75em}
We instantiate \modelname as an iterative algorithm over a dynamically maintained hypothesis set. At each iteration, candidate mechanisms are sampled from a distribution conditioned on the current state, and the next experiment is selected by maximizing an inter-hypothesis disagreement objective over this set. The resulting observation is incorporated by refitting each candidate hypothesis on the augmented dataset, computing its empirical loss, and applying stability-based filtering to retain consistent mechanisms and eliminate unstable ones.



\vspace{-0.5em}
\subsection{Problem Formulation}
\vspace{-.25em}

\label{sec:problem_formulation}
\begin{wrapfigure}{r}{0.5\textwidth}
\vspace{-3.em}
\begin{minipage}{\linewidth}
\begin{algorithm}[H]
\caption{\modelname}
\label{alg:mainalg}
\begin{algorithmic}[1]

\Require Oracle $\mathcal{O}$, Dataset $\mathcal{D}$, Budget $B$, State $\mathcal{S}_t$, Search Region $\mathcal{R}$, Memory $\mathcal{E}$, LLM $\pi_\theta$, Hypothesis Set $\mathcal{H}_t$, Confidence Threshold $\tau_{\rm conf}$, Confidence Score $c_t$  
\Statex \textcolor{gray}{{\# {Initialize data and experience buffer}}}

\State $\mathcal{D}_0, c_0, \mathcal{E}_0 \gets \emptyset, \emptyset, \texttt{InitMemory}()$

\For{$t = 0,\ldots,B-1$}
    \State $S_t \gets (\mathcal{D}_t, \mathcal{E}_t, \mathcal{H}_t)$
    \Statex \textcolor{gray}{{\# {Propose hypotheses and search regions}}}
    \State $\mathcal{H}_t, \mathcal{R}_t  \gets \texttt{GenHyp}(\pi_\theta^{\rm large},\pi_\theta^{\rm small},S_t)$
\Statex \textcolor{gray}{{\# {Select acquisition mode}}}

    \If{$c_t < \tau_{\rm conf}$}
    \State $\texttt{mode} \gets \texttt{Disambiguate}$
    \State $\Delta_t \gets \texttt{Disagree}(\mathcal{H}_t,\mathcal{D}_t)$
    \Else
    \State $\texttt{mode} \gets \texttt{Refine}$
    \State $\Delta_t \gets \emptyset$
    \EndIf

\Statex \textcolor{gray}{{\# Acquire new experiment}}
    \State $\mathbf{x}_{t+1} \gets \texttt{Acquire}(\mathcal{D}_t,\mathcal{H}_t,\mathcal{R}_t,\texttt{mode}, \Delta_t)$
    \State $y_{t+1} \gets \mathcal{O}(\mathbf{x}_{t+1})$
    \State $\mathcal{D}_{t+1} \gets
    \mathcal{D}_t \cup \{(\mathbf{x}_{t+1},y_{t+1})\}$

\Statex \textcolor{gray}{{\# Refine and update memory}}

    \State $\hat{m}_{t+1},c_{t+1} \gets \texttt{RefineHyp}(\mathcal{D}_{t+1},\mathcal{H}_t)$
    \State $\tilde{m}_{t+1},\tilde{c}_{t+1}
    \gets \texttt{ConfGate}(\hat{m}_{t+1},c_{t+1})$
    \State $\mathcal{E}_{t+1} \gets
    \texttt{UpdateMemory}()$
\EndFor

\State \Return $\tilde{m}_B$

\end{algorithmic}
\end{algorithm}
\end{minipage}
\vspace{-1em}
\end{wrapfigure} 
We frame scientific discovery as an active experimental design task, optimizing hypothesis selection under a fixed resource budget. Let $\mathcal{M}$ denote a space of candidate mechanisms, where each mechanism $m \in \mathcal {M} $ defines a predictive mapping $f_m: \mathcal{X} \rightarrow \mathcal{Y}$. The objective is to recover the unknown ground-truth mechanism $m^\star \in \mathcal{M}$. At each round $t$, the learner selects an experiment $x_t \in \mathcal{X}$ that yields an observation $y \sim p(\cdot \mid \bx, m^\star)$, where $p$ is the observation model. Unlike static settings, where the data are fixed a priori, this task requires active data acquisition. Consequently, after $t$ rounds, the accumulated dataset  $\calD_t = \{(\bx_i, y_i)\}_{i=1}^t$ grows iteratively until the total experimental budget $B$ is exhausted. At each step, the process maintains a discovery state $\calS_t = (\calD_t, \calE_t, \calH_t)$, where $\calE_t$ is structured memory summarizing prior hypotheses and evidence, and $\calH_t$ is the current set of plausible mechanisms given as: $\mathcal H_t = \{m^{(k)}\}_{k=1}^K \subseteq \mathcal{M}$. {A discovery policy $\pi$ maps the current state to the next experiment, producing a sequence of adaptive queries $\bx_{t+1} = \pi_t(\calS_t)$, observations $y_{t+1} \sim p(\cdot \mid \bx_{t+1}, m^\star)$, and updated states $\mathcal S_{t+1}$. The objective is to produce a final estimate $\hat{m}_B$ that minimizes expected mechanism error $\E[\mathcal{L}(\hat{m}_B, m^\star)]$, where $\mathcal L$ is a domain-dependent loss, and the expectation is over trajectories induced by $\pi$ and $p$, with a fixed $m^\star$.



\vspace{-0.75em}
\subsection{Hypothesis Generation}
\vspace{-.25em}

To mitigate the path-dependence and premature collapse of one-shot LLM hypothesis generation under sparse data~\citep{chen2025hypospace}, \modelname decouples \textit{hypothesis diversity} from \textit{hypothesis synthesis}. As shown in Algorithm~\ref{alg:mainalg}, $\texttt{GenHyp}(\cdot)$ decouples exploration from synthesis via asymmetric model roles. A smaller LLM $\pi_\theta^{\rm small}$ is sampled in batches to generate candidate hypotheses conditioned on the current state $\mathcal{S}_t$. These are grouped into structural mechanism families, and sampling continues until the distribution stabilizes, yielding a hypothesis set $\mathcal{H}_t$. A larger LLM $\pi_\theta^{\rm large}$ then conditions on $\mathcal{H}_t$ to produce a structured proposal containing a primary hypothesis, alternative hypotheses, and diagnostic search regions $\mathcal{R}_t \subseteq \Theta$. Thus, LLMs are used to define a hypothesis space for data acquisition via experimentation, rather than to produce a final answer.

\vspace{-0.75em}

\subsection{Hypothesis-Conditioned Experiment Selection}
\vspace{-.25em}
The data acquisition policy is governed by the disagreement of the hypotheses $\mathcal {H} _ t$ within the LLM-generated search region $\mathcal{R}_t$. Given the confidence score $c_t$, \modelname first determines whether the current hypothesis is sufficiently stable to serve as the basis for local refinement. If $c_t < \tau_{\rm conf}$, \modelname retains the full hypothesis set $\mathcal{H}_t$ and selects experiments for mechanism disambiguation; otherwise, it treats the current mechanism as stable and shifts acquisition to refinement. In \texttt{Disambiguate} mode \modelname computes $\Delta_t$ = \texttt{Disagree}$(\mathcal{H}_t,\mathcal{D}_t),$ which governs the acquisition strategy and \texttt{Acquire($\cdot$)} selects $x_{t+1} \in \mathcal{R}$ where candidate mechanisms in $\mathcal{H}_t$ make maximally divergent predictions. In \texttt{Refine} mode, \texttt{Acquire($\cdot$)} instead selects experiments that improve the fit or parameterization of the supported mechanism family. Unlike Bayesian experimental design, which typically optimizes information gain under a predefined probabilistic model class, \modelname constructs and revises explicit mechanism hypotheses and uses their predicted disagreements to guide experiment selection. This makes acquisition depend on the status of the evolving hypothesis space, rather than predictive improvement alone.


\vspace{-0.75em}
\subsection{Hypothesis Optimization and Confidence Feedback}
\vspace{-0.5em}
After each experiment, the new observations from $\mathcal{O}$ are incorporated into $\mathcal{D}_{t}$ to produce the new dataset $\mathcal{D}_{t+1}$. As shown in Algorithm~\ref{alg:mainalg}, $\texttt{RefineHyp}(\cdot)$ fits $\mathcal{H}_t$ to the new dataset producing a refined mechanism $\hat{m}_{t+1}$ along with a confidence score $c_{t+1}$. This step turns each generated hypothesis structure into a data-evaluated mechanism on $\mathcal{D}_{t+1}$.
To assess robustness under adaptive data collection, we introduce a confidence gate applied via $\texttt{ConfGate}(\cdot)$. Since $\mathcal{D}_{t+1}$ may be biased toward the currently selected regions, this step performs bootstrap resampling and refits the candidate mechanism across these datasets, measuring agreement across the resulting fits. Consistent hypotheses are assigned higher confidence $\tilde{c}_{t+1}$, while those exhibiting instability in structure or predictions are treated as brittle. The confidence-adjusted mechanism $\tilde{m}_{t+1}$ is then written back into memory through $\texttt{UpdateMemory}(\cdot)$, informing subsequent hypothesis generation and acquisition decisions. Appendix~\ref{appendix_autoscilab_method} provides a complete algorithmic specification and implementation details.

\vspace{-0.75em}
\subsection{Implementation Details}
\vspace{-0.5em}
We use \texttt{GPT-4o-mini} as the primary LLM backbone and \texttt{Qwen/Qwen2.5-7B-Instruct} for the smaller models used in adaptive ensembling. Data acquisition follows the NewtonBench setup~\citep {zheng2026newtonbench}, where the oracle is a noiseless black-box function $u \mapsto f_{\mathrm{target}}(u)$ defined over an open set $U$ of achievable target-input values. For equation discovery, the refinement backend uses \texttt{PySR} with 800 iterations per fitting call, together with direct numerical fitting of candidate mechanism skeletons when available. For graph discovery, refinement uses \texttt{BFGS} with 800 iterations. All models are used in inference mode without task-specific finetuning; additional implementation details and hyperparameters are reported in Appendix~\ref{appendix_baselines}.

\vspace{-1.em}
\section{\benchmark: Benchmark for Active Scientific Discovery}
\vspace{-0.6em}
Real-world scientific discovery requires not only inferring governing laws from observations, but also choosing experiments that yield the most informative data. Existing benchmarks reduce discovery to passive inference on fixed datasets, bypassing the experimental-design problem that is critical when observations are costly. \textcolor{black}{To address this gap, we introduce \textbf{\benchmark} a two-dataset benchmark suite for active, closed-loop scientific discovery (Figure~\ref{fig:main_benchmark})}, each based on \emph{physically grounded laws} and a queryable experimental system where the underlying law and parameters are hidden, relevant variables are unknown a priori, and discovery must occur within a fixed experimental budget.
\vspace{-.75em}
\subsection{\chem: Active Enzyme-Kinetic Law Discovery}
\vspace{-0.25em}
\paragraph{Task Formulation.}
Enzyme-kinetic rate laws describe how reaction rates vary as a function of experimental conditions. In \chem (Figure~\ref{fig:main_benchmark}~(a)), each task simulates an enzyme-catalyzed reaction governed by a hidden kinetic mechanism and hidden parameters; the learner must recover the symbolic rate law from budget-limited experiments. Each experiment is specified via a shared 7-dimensional interface for substrate, inhibitor, second-substrate, and product concentrations, enzyme loading, temperature, and pH, and returns the observed initial rate $r_0$ along with auxiliary mass-balance observables. The true rate law, including its functional form and which inputs actually appear, is withheld from the learner throughout. Candidate mechanisms can produce indistinguishable behavior over restricted regions of the design space, and the correct law is often recoverable only through experiments that deliberately isolate individual dependencies.
\vspace{-1em}

\paragraph{Dataset Construction.}
The benchmark is drawn from a curated, mechanistically grounded hypothesis space comprising standard kinetic families, structured compositions thereof, and extended mechanisms beyond the standard textbook library, yielding 57 curated tasks in total. We report results across three complexity tiers: \textit{Easy} (standard families: Michaelis-Menten, competitive inhibition), \textit{Medium} (structured compositions: mixed inhibition, substrate inhibition), and \textit{Hard} (extended mechanisms: cooperative binding, allosteric regulation). 

\begin{figure}[!htbp]
    \centering
    \includegraphics[width=0.98\linewidth]{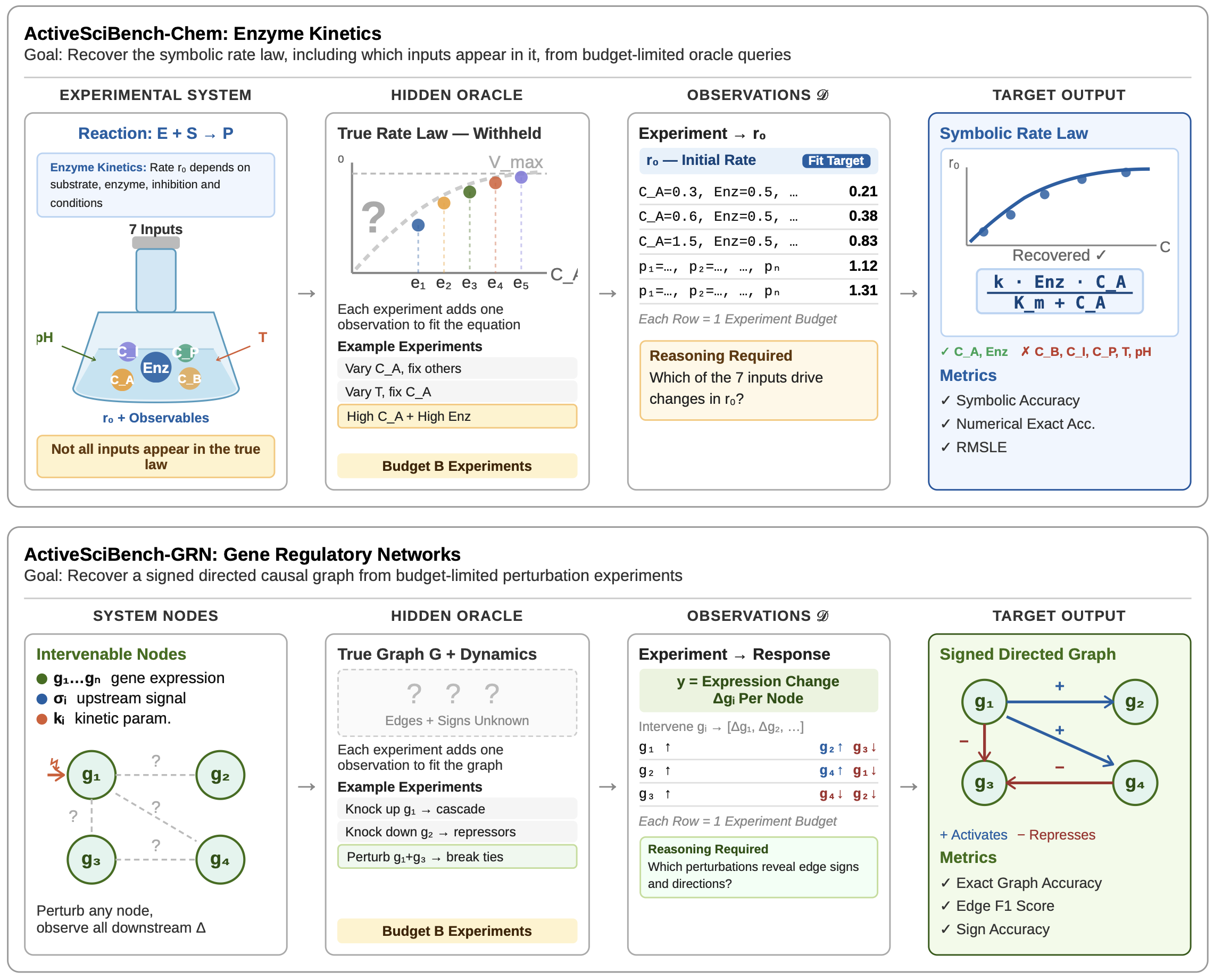}
    \caption{\small {\textbf{Overview of \benchmark}:} (a) \textbf{\chem:} Symbolic enzyme rate law recovery; (b) \textbf{\grn:} Signed directed gene regulatory graph inference.}
    \label{fig:main_benchmark}
    \vspace{-1.25em}
\end{figure}

\subsection{\grn: Active Causal Graph Discovery}
\vspace{-.5em}
\paragraph{Task Formulation.}
Gene regulatory networks are signed, directed graphs describing which genes or regulators activate or repress other genes. In \grn~(Figure~\ref{fig:main_benchmark}~(b)), each task simulates a hidden regulatory system with an unknown graph structure and nonlinear dynamics, aiming to recover the causal graph from within an experimental budget. The system consists of intervenable gene expression nodes, upstream signals, and kinetic parameters. Unlike \chem, \grn operates in a discrete intervention space: each experiment knocks up, knocks down, or perturbs specific nodes, and the learner observes all downstream expression changes. The true graph, including edge presence, direction, and sign (activation vs.\ repression) along with the governing nonlinear dynamics, is withheld from the learner throughout. Different motifs can produce similar observations under weak interventions, and hidden parameters can make the same motif look qualitatively different across tasks. The learner must therefore jointly identify topology, sign, and effective dynamics from sparse data, requiring informative perturbation choices rather than response-surface fitting. The core challenge is inferring which variables are relevant, but here the discovery target is a structured causal graph rather than a symbolic equation, extending the benchmark suite to broader coverage of scientific discovery problems.
\vspace{-1.0em}
\paragraph{Dataset Construction.}
The benchmark is curated from a small set of canonical regulatory motifs spanning increasing structural complexity, from feedforward activation to repression, feedback, and switching behavior. It contains 5 motif families, each instantiated at 3 native difficulty levels and 3 topological versions, yielding 45 tasks per random seed with the capacity to generate more. The data split in \grn corresponds directly to the difficulty levels, reflecting progressively sharper and more nonlinear parameter regimes. \textit{Easy} tasks exhibit quasi-linear responses where small perturbations reveal the graph clearly; \textit{Medium} tasks introduce nonlinearity where saturation effects obscure weak edges; and \textit{Hard} tasks feature bistability and switching behavior, where the graph is identifiable only through carefully designed multi-node perturbations. Appendix~\ref{appendix_benchmark_details} details the construction details, family definitions, filtering rules, and tasks.
\vspace{-0.5em}

\begin{table*}[!htbp]
\centering
\vspace{-1.5em}
\caption{\small \textbf{Quantitative performance comparison} of baselines across all benchmarks. \textit{ED} denotes whether a method incorporates experiment design. Metrics for \textbf{NewtonBench} and \textbf{\chem}: SA = Symbolic Accuracy (\%), Ex. = Exact Accuracy (\%), RMSLE = Root Mean Squared Log Error. \textbf{\grn}: F1 = Edge F1 (\%), Ex. = Exact Graph Accuracy (\%), Sign = Sign Accuracy (\%).}
\vspace{-0.5em}
\label{tab:main_results}
\fontsize{10}{10}\selectfont
\renewcommand{\arraystretch}{1.05}
\resizebox{\textwidth}{!}{
\begin{tabular}{lc ccc ccc ccc ccc}
\toprule
\multirow{2}{*}{\textbf{Method}} & \multirow{2}{*}{\textbf{ED}} 
& \multicolumn{3}{c}{\textbf{Easy}} 
& \multicolumn{3}{c}{\textbf{Medium}} 
& \multicolumn{3}{c}{\textbf{Hard}} 
& \multicolumn{3}{c}{\textbf{Overall}} \\
\cmidrule(lr){3-5}
\cmidrule(lr){6-8}
\cmidrule(lr){9-11}
\cmidrule(lr){12-14}
&
& \textbf{SA}($\uparrow$)  & \textbf{Ex.}($\uparrow$)  & \textbf{RMSLE}($\downarrow$)
& \textbf{SA}($\uparrow$) & \textbf{Ex.}($\uparrow$)  & \textbf{RMSLE}($\downarrow$)
& \textbf{SA}($\uparrow$) & \textbf{Ex.}($\uparrow$)  & \textbf{RMSLE}($\downarrow$)
& \textbf{SA}($\uparrow$) & \textbf{Ex.}($\uparrow$)  & \textbf{RMSLE}($\downarrow$) \\
\midrule

\multicolumn{14}{c}{\textit{\textbf{NewtonBench}}} \\[2pt]
PySR
    & \texttimes 
    & 38.89              & \underline{88.89} & \underline{0.030}
    & \underline{27.78} & \underline{75.00} & \underline{0.125}
    & \underline{5.56}  & \underline{59.72} & \textbf{0.398}
    & 24.07              & \underline{74.54} & \underline{0.182} \\
Bayesian Optimization
    & \checkmark 
    & \underline{40.28} & 84.72 & 0.070
    & \underline{27.78} & 70.83 & 0.212
    & \underline{5.56}  & 50.00 & 0.730
    & \underline{24.54} & 68.52 & 0.334 \\
Bayesian Experimental Design
    & \checkmark 
    & 19.44 & 78.57 & 0.307
    & 11.11 & 66.18 & 0.449
    & 2.78  & 45.31 & 1.010
    & 11.11 & 63.86 & 0.577 \\
LLM-only
    & \checkmark 
    & 16.67 & 19.44 & 3.572
    & 2.78  & 2.78  & 4.738
    & 0.00  & 0.00  & 6.757
    & 6.48  & 7.41  & 5.039 \\
Code-assisted LLM
    & \checkmark 
    & 13.89 & 19.44 & 4.372
    & 5.56  & 11.11 & 5.437
    & 2.78  & 0.00  & 4.897
    & 7.41  & 10.19 & 4.912 \\
\rowcolor{gray!10}
\textbf{\modelname\ (Ours)}
    & \checkmark 
    & \textbf{79.20} & \textbf{93.10} & \textbf{0.018}
    & \textbf{72.20} & \textbf{84.70} & \textbf{0.040}
    & \textbf{51.40} & \textbf{66.70} & \underline{0.404}
    & \textbf{67.60} & \textbf{81.50} & \textbf{0.150} \\

\midrule

\multicolumn{14}{c}{\textit{\textbf{\chem}}} \\[2pt]
PySR
    & \texttimes 
    & {33.33} & {33.33} & 0.455
    & 0.00               & {11.54} & 1.545
    & 0.00               & 0.00              & {0.601}
    & {7.89}  & {15.79} & 1.212 \\
Bayesian Optimization
    & \checkmark 
    & 22.22 & 14.29 & 2.001
    & 0.00  & 3.85  & 1.338
    & 0.00  & 0.00  & 6.564
    & 5.26  & 6.98  & 1.960 \\
Bayesian Experimental Design
    & \checkmark 
    & \textbf{77.78} & \underline{77.78} & \textbf{0.028}
    & \underline{19.23} & \underline{30.77} & \underline{0.333}
    & 0.00               & 0.00              & \underline{0.187}
    & \underline{31.58} & \underline{39.47} & \underline{0.249} \\
LLM-only
    & \checkmark 
    & 0.00 & 0.00 & 0.563
    & 0.00 & 0.00 & 0.777
    & 0.00 & 0.00 & 0.849
    & 0.00 & 0.00 & 0.764 \\
Code-assisted LLM
    & \checkmark 
    & 0.00 & 0.00 & 0.554
    & 0.00 & 0.00 & 0.869
    & 0.00 & 0.00 & 0.894
    & 0.00 & 0.00 & 0.824 \\
\rowcolor{gray!10}
\textbf{\modelname\ (Ours)}
    & \checkmark 
    & \underline{55.56} & \textbf{88.89} & \underline{0.298}
    & \textbf{22.22} & \textbf{37.04} & \textbf{0.179}
    & \textbf{42.86} & \textbf{52.38} & \textbf{0.154}
    & \textbf{35.09} & \textbf{50.88} & \textbf{0.189} \\

\midrule
&
& \textbf{F1}($\uparrow$)  & \textbf{Ex.}($\uparrow$)  & \textbf{Sign}($\uparrow$)
& \textbf{F1}($\uparrow$)  & \textbf{Ex.}($\uparrow$)  & \textbf{Sign}($\uparrow$)
& \textbf{F1}($\uparrow$) & \textbf{Ex.}($\uparrow$)  & \textbf{Sign}($\uparrow$)
& \textbf{F1}($\uparrow$)  & \textbf{Ex.}($\uparrow$)  & \textbf{Sign}($\uparrow$) \\
\midrule

\multicolumn{14}{c}{\textit{\textbf{\grn}}} \\[2pt]
GENIE3
    & \texttimes 
    & 41.54              & 0.00               & 85.33
    & 35.00              & 0.00               & 71.89
    & {39.27} & 0.00               & 80.00
    & 38.60              & 0.00               & 79.07 \\
GIES
    & \texttimes 
    & \textbf{70.12} & \underline{20.00} & \underline{97.78}
    & {47.07} & 0.00               & {80.56}
    & 51.61              & 0.00               & 79.00
    & \underline{56.27} & \underline{6.67}  & {85.78} \\
NOTEARS
    & \texttimes 
    & 31.52 & 0.00               & 66.67
    & 25.71 & 0.00               & 56.67
    & 25.57 & \underline{6.67} & 51.11
    & 27.60 & 2.22               & 58.15 \\
Random sampling
    & \checkmark 
    & 33.64 & 0.00               & 86.67
    & 33.46 & 0.00               & \underline{86.67}
    & 39.58 & \underline{6.67} & 80.00
    & 35.56 & 2.22               & 84.44 \\
Uncertainty sampling
    & \checkmark 
    & {55.26} & {13.33} & 93.33
    & \underline{49.63}              & 0.00               & \underline{86.67}
    & 45.40              & 0.00               & \underline{84.44}
    & 50.10              & {4.44}  & \underline{88.15} \\
LLM-only
    & \checkmark
    & 53.61 & 0.00 & 81.67
    & 46.11 & 0.00 & 77.78
    & 51.51 & 0.00 & 78.89
    & 50.41 & 0.00 & 79.44 \\
Code-assisted LLM
    & \checkmark
    & 51.73 & 0.00 & 82.78
    & 57.65 & 0.00 & 91.67
    & \underline{54.63} & 0.00 & 83.33
    & 54.67 & 0.00 & 85.98 \\

\rowcolor{gray!10}
\textbf{\modelname\ (Ours)}
    & \checkmark 
    & \underline{66.27} & \textbf{26.67} & \textbf{100.0}
    & \textbf{83.13} & \textbf{40.00}  & \textbf{100.0}
    & \textbf{68.06} & \textbf{26.67} & \textbf{94.44}
    & \textbf{72.49} & \textbf{31.11} & \textbf{98.15} \\

\bottomrule
\end{tabular}
}
\vspace{-1.25em}
\end{table*}
\section{Experiments}
\label{sec:experiments}
\vspace{-0.65em}
\subsection{Experimental Setup}
\vspace{-0.65em}

\paragraph{Datasets.}We evaluate \modelname in a closed-loop scientific discovery setting where the system iteratively designs experiments and refines hypotheses under a fixed oracle query budget. Our study includes both quantitative comparisons with prior methods and targeted ablations. Specifically, we conduct experiments on: \textbf{NewtonBench}, spanning 12 physics domains across multiple difficulty levels and variants; \textbf{\chem}, a suite of compositional enzyme kinetics tasks; and \textbf{\grn}, which focuses on graph-structured discovery in gene regulatory networks. For \chem and NewtonBench, we report: (i) symbolic accuracy, (ii) predictive error via RMSLE, (iii) numerical exact accuracy. Symbolic accuracy is stricter than numerical accuracy, as approximate fits may not recover the true form. For \grn (structure discovery), we evaluate structural recovery using edge-level precision, recall, F1, and sign accuracy (activation vs.\ repression), and mechanistic recovery via exact graph accuracy and motif accuracy. Additional metric details are provided in Appendix~\ref{appendix_metrics}.
\vspace{-0.75em}

\paragraph{Baselines.}We compare against a broad set of baselines: \textbf{Symbolic regression methods} such as PySR~\citep{cranmer2023interpretable} on fixed datasets without experiment design and \textbf{Active learning methods} (Bayesian Optimization, Bayesian Experimental Design) that select experiments but do not model symbolic structure. We further experiment with LLM-only prompting and code-assisted LLMs. For \grn, we additionally evaluate \textbf{Graph Discovery Baselines} GENIE3 \cite{genie}, GIES \cite{gies}, and NOTEARS \cite{tears} (offline), as well as Random and Uncertainty sampling. For ablations and diagnostic analyses, we use stratified, representative subsets to improve computational tractability (Appendix~\ref{appendix_experimental_protocol}).  For all LLM-based methods, we use \texttt{GPT-4o-mini} as the primary model with \texttt{Qwen2.5-7B-Instruct} as the smaller local ensemble model. Appendix~\ref{app:stats} reports results with standard deviations.
\vspace{-1.em}
\subsection{Main Results}
\vspace{-0.5em}
As shown in Table~\ref{tab:main_results}, we evaluate \modelname using \texttt{GPT-4o-mini} across NewtonBench, \chem, and \grn under fixed oracle budgets. Across all three benchmarks, \modelname achieves the strongest overall recovery, but the source of advantage differs by setting: NewtonBench tests symbolic identifiability with known variables, \chem requires discovering relevant kinetic variables, and \grn requires recovering graph structure from sparse interventions. Appendix~\ref{app:model_capability} contains experiments using \texttt{Qwen-3} family LLMs.
\vspace{-2em}

\paragraph{NewtonBench.}
NewtonBench isolates symbolic recovery when relevant variables are known. Strong fitting baselines such as PySR and Bayesian Optimization achieve high exact accuracy ($74.54\%$, $68.52\%$) but low symbolic accuracy ($24.07\%$, $24.54\%$), indicating overfitting to observed regimes without recovering the underlying law. LLM-based methods perform poorly overall ($<8\%$ SA). In contrast, \modelname{} achieves $67.60\%$ symbolic accuracy and $81.50\%$ exact accuracy, substantially outperforming all baselines. This gap highlights that \textit{hypothesis-conditioned experimentation improves structural identification rather than merely numerical fit}.
\vspace{-1em}

\paragraph{\chem.}
\chem requires identifying both relevant variables and kinetic structure through interaction with the experimental interface. BED is the strongest baseline ($31.58\%$ SA), performing well on easy tasks ($77.78\%$ SA) but collapsing on hard settings ($0.00\%$ SA/Ex.), suggesting limited coverage beyond the candidate model class. We observe that LLM-only and code-assisted LLM baselines consistently achieve $0.00\%$ SA/Ex. Although they often produce plausible rate laws, they tend to default to generic textbook templates rather than testing variable relevance or distinguishing kinetic families. Thus, their outputs can be locally reasonable while failing strict symbolic-equivalence and exact-recovery criteria. In contrast, \modelname achieves the best overall performance ($35.09\%$ SA, $50.88\%$ Ex.) and remains robust on hard tasks ($42.86\%$ SA, $52.38\%$ Ex.). The results suggest that \textit{LLM-generated hypotheses enable exploration beyond fixed mechanism libraries, which is critical for nonstandard kinetics}.
\vspace{-1.em}
\paragraph{\grn.}
\grn evaluates graph-structured mechanism recovery. Offline methods recover partial structure but rarely the full graph: GIES achieves $56.27\%$ F1 but only $6.67\%$ exact accuracy. Active baselines improve edge recovery (e.g., uncertainty sampling: $50.10\%$ F1) but still fail to recover full graphs ($4.44\%$ Ex.). \modelname{} significantly outperforms all baselines, achieving $72.49\%$ F1, $31.11\%$ exact graph accuracy, and $98.15\%$ sign accuracy. We observe that most methods perform very well on the sign metric, indicating that it is easier to determine whether a node suppresses, activates, or has no effect on other nodes. This demonstrates that \textit{accurate graph recovery requires targeted, hypothesis-driven perturbations that disambiguate competing signed structures}, rather than passive fitting or uncertainty-based sampling.

\vspace{-0.5em}

\vspace{-0.25em}
\section{Analysis}
\vspace{-0.5em}
\subsection{Qualitative Analysis}
\vspace{-0.55em}
\begin{wrapfigure}{r}{0.45\linewidth}
    \vspace{-4.3em}
    \centering
    \includegraphics[width=\linewidth]{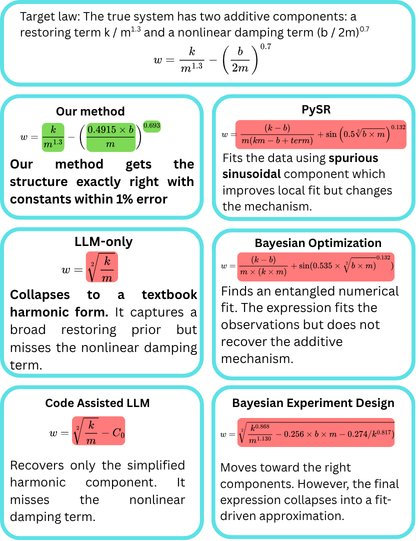}
    \vspace{-1.5em}
    \caption{ \small
    \textbf{Qualitative NewtonBench case study.}
    \modelname recovers the correct symbolic structure, while other baselines introduce spurious terms, collapse to incorrect families, or recover only simplified harmonic forms.
    }
    \label{fig:qualitative_newtonbench_case}
    \vspace{-1.0em}
\end{wrapfigure}

Figure~\ref{fig:qualitative_newtonbench_case} presents a qualitative NewtonBench case study illustrating failure modes under limited-budget mechanism discovery. The target system contains two additive components: a restoring force term and a nonlinear damping term. Fit-driven baselines such as PySR and Bayesian optimization achieve low local error by introducing spurious or entangled terms, but fail to recover the correct mechanistic structure. In particular, PySR inserts an unnecessary sinusoidal component, while Bayesian optimization converges to numerically accurate but mechanistically incorrect expressions. LLM-only and code-assisted LLM methods instead collapse to simplified textbook harmonic forms, identifying only partial structure and missing the nonlinear damping behavior. Bayesian experiment design moves toward the correct components, but ultimately still converges to a fit-driven approximation. In contrast, \modelname recovers the correct additive structure and closely matches the hidden constants, including a recovered coefficient of $0.4915 \approx 1/2$ and a damping exponent within roughly $1\%$ of the ground truth, highlighting the importance of discriminative experimentation in accurate mechanism recovery.

\vspace{-0.5em}

\subsection{Ablation Study}
\vspace{-0.5em}
Figure~\ref{fig:ablation_study} reports single-component removal results across all three benchmarks. Removing hypothesis-conditioned acquisition causes a substantial drop across all benchmarks, showing that candidate mechanisms must guide data acquisition under limited oracle budgets. Other components contribute differently depending on the source of difficulty. NewtonBench stresses sparse functional identification through counterfactual laws, making diverse hypothesis generation and mechanism stability important; without them, the LLM is drawn toward canonical physics laws. \chem requires reasoning over the input space and identifying mechanistically relevant variables, making memory crucial for disambiguating competing mechanisms. \grn relies on accumulating perturbation evidence, making evidence preservation and intervention selection more central. These ablation patterns also support the diversity of our suite: NewtonBench, \chem, and \grn are sensitive to different removed components, suggesting they probe complementary discovery capabilities rather than one shared fitting problem. Overall, the results show that \modelname functions as a closed-loop pipeline across discovery settings, with each component contributing meaningfully.
\begin{figure}[!htbp]
    \vspace{-0.75em}
    \centering
    \includegraphics[width=0.99\linewidth]
    {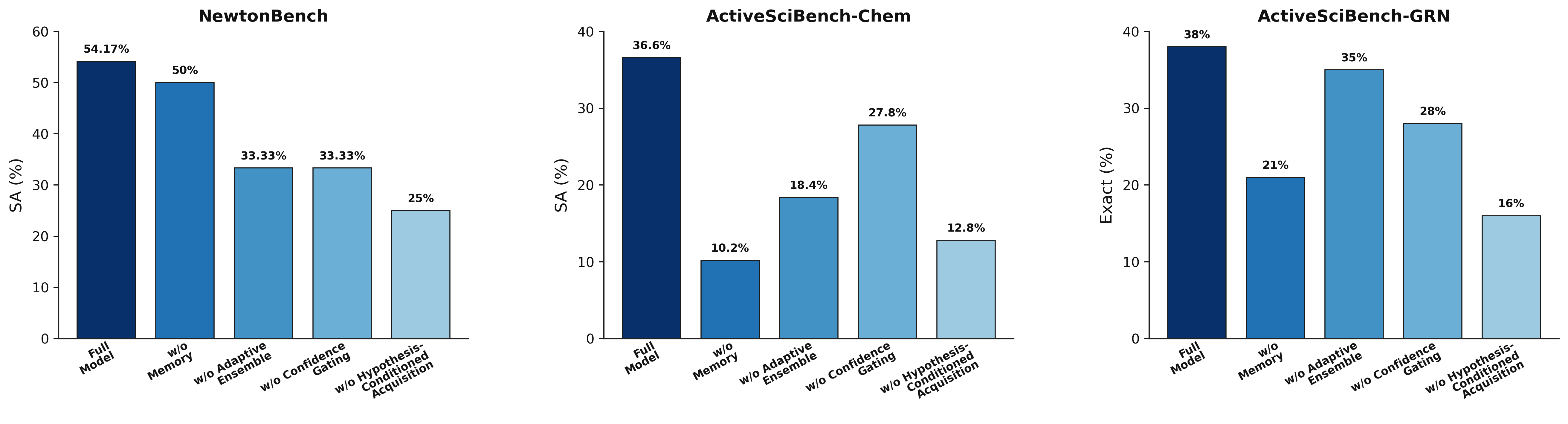}
    \vspace{-1.em}
    \caption{\text{Ablation study across all benchmarks} removing one component from \modelname.
}
    \vspace{-1.5em}
    \label{fig:ablation_study}
\end{figure}

\subsection{Experiment Budget Analysis}
\vspace{-0.5em}
Figure~\ref{fig:budget_ablation_study} plots recovery versus query budget. On NewtonBench and \grn, the second-best baseline     fails to match \modelname{}'s fixed-budget performance even at $5\times$ the query count. For \chem, BED closes the low-budget gap only by budget 60, using three times as many queries as the $B=20$ setting of \modelname{}. Separately, we also measure sample efficiency as the query budget each baseline needs to match \modelname{}'s fixed-budget performance. The strongest active baselines require substantially more queries: $2.60$--$3.10\times$ on NewtonBench, $2.33$--$2.47\times$ on \chem, and $3.90$--$4.60\times$ on \grn; LLM-only and code-assisted variants require $5.20$--$14.40\times$ more queries depending on the benchmark (Appendix~\ref{relative_query_cost}).
By conditioning acquisition on competing hypotheses, each oracle call is more likely to resolve structural ambiguity across symbolic laws, kinetic rate mechanisms, and signed regulatory graphs. Together, these results show that \textit{\modelname{} is significantly more sample efficient than the baselines}.

\begin{figure}[!htbp]
    \vspace{-.5em}
    \centering
    \includegraphics[width=0.9\linewidth]{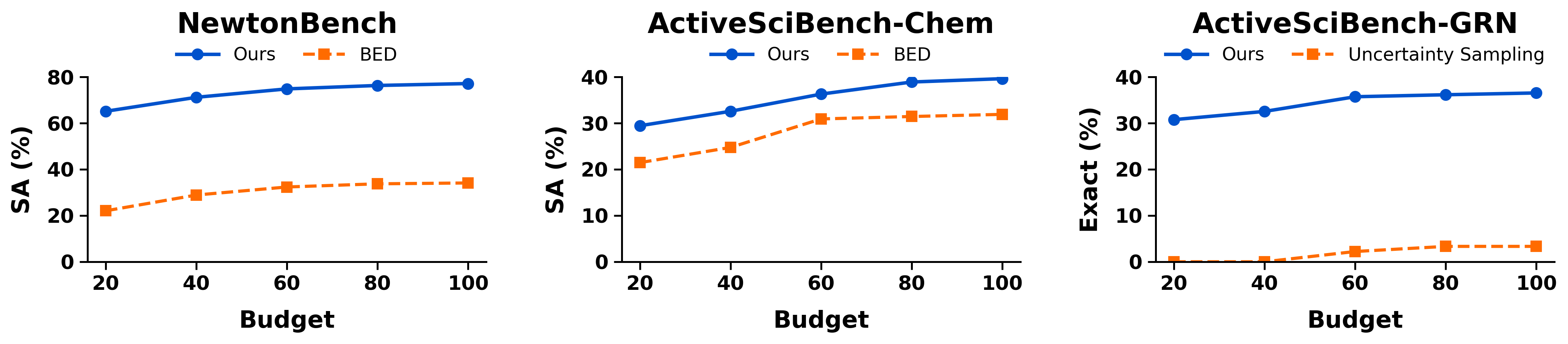}
    \vspace{-0.5em}
    \caption{\text{Budget ablations across benchmarks} showing the recovery metric versus query budget.
    }
    \label{fig:budget_ablation_study}
    \vspace{-.5em}
\end{figure}


\vspace{-0.5em}
\section{Conclusion}
\label{sec:conclusion}
\vspace{-0.5em}
We introduced \modelname, a closed-loop algorithm for scientific discovery that formalizes discovery as iterative experimental design over an evolving hypothesis set. \modelname instantiates a structured algorithmic loop that (i) \textit{generates a diverse hypothesis set}, (ii) selects experiments via a \textit{hypothesis-conditioned acquisition objective}, and (iii) refines candidates through\textit{ data-driven optimization with feedback}. This formulation shifts the objective of data acquisition from predictive uncertainty reduction to hypothesis discrimination, improving true mechanism recovery under limited experimental budgets. To overcome the lack of evaluation settings for closed-loop discovery, we introduce \textbf{\chem} and \textbf{\grn}, \textit{benchmarks that recast scientific discovery as an active, budget-constrained process requiring joint experiment design, variable selection, and recovery of underlying mechanisms}, enabling systematic evaluation beyond static function fitting. Across NewtonBench, \chem, and \grn, \modelname achieves higher symbolic and structural recovery with fewer queries than prior methods, demonstrating that \textit{aligning data acquisition with hypothesis discrimination rather than predictive accuracy improves both efficiency and reliability of scientific discovery}.

\vspace{-0.15in}
\paragraph{Limitations.}
\modelname uses simulator-based oracles, so physical-lab noise, failures, costs, and operational constraints are not fully modeled. Performance also depends on the quality of LLM-generated hypotheses and on the coverage of the parser and refinement backends. Broader domains, richer refinement tools, and real-world validation remain future work.


\bibliographystyle{plain}
\bibliography{neurips_2026}

\appendix
\subsection*{Reproducibility Statement}
To ensure reproducibility, we provide the relevant implementation and experimental details throughout the paper, including the overall methodology described in Section~\ref{sec:method} and Appendix~\ref{appendix_autoscilab_method}, and the LLM prompts listed in Appendix~\ref{app:prompts}. We also document the datasets used in our experiments in Appendix~\ref{appendix_benchmark_details} and release the accompanying code and data to support future research.

\subsection*{Impact Statement}
LLM-AutoSciLab accelerates scientific discovery by automating hypothesis-driven experimentation, with potential benefits for researchers in biology, chemistry, and physics who face costly experimental budgets, reducing the number of experiments needed to recover governing mechanisms and lowering barriers for smaller research groups. The primary risks are over-reliance on model outputs in safety-critical domains such as drug discovery, where a plausible but incorrect mechanistic law could have downstream consequences, and the inheritance of LLM biases that may systematically favor well-represented mechanisms over genuinely novel ones. The framework currently targets simulator-based discovery rather than direct laboratory deployment, limiting immediate risk, but domain-specific safety review remains essential before application in sensitive real-world contexts. We do not anticipate direct dual-use concerns.

\section*{Appendix}
\section{Additional Results}
\subsection{Noise Sensitivity}
\begin{figure}[H]
    \centering

    \begin{subfigure}[t]{0.32\linewidth}
        \centering
        \includegraphics[width=\linewidth]{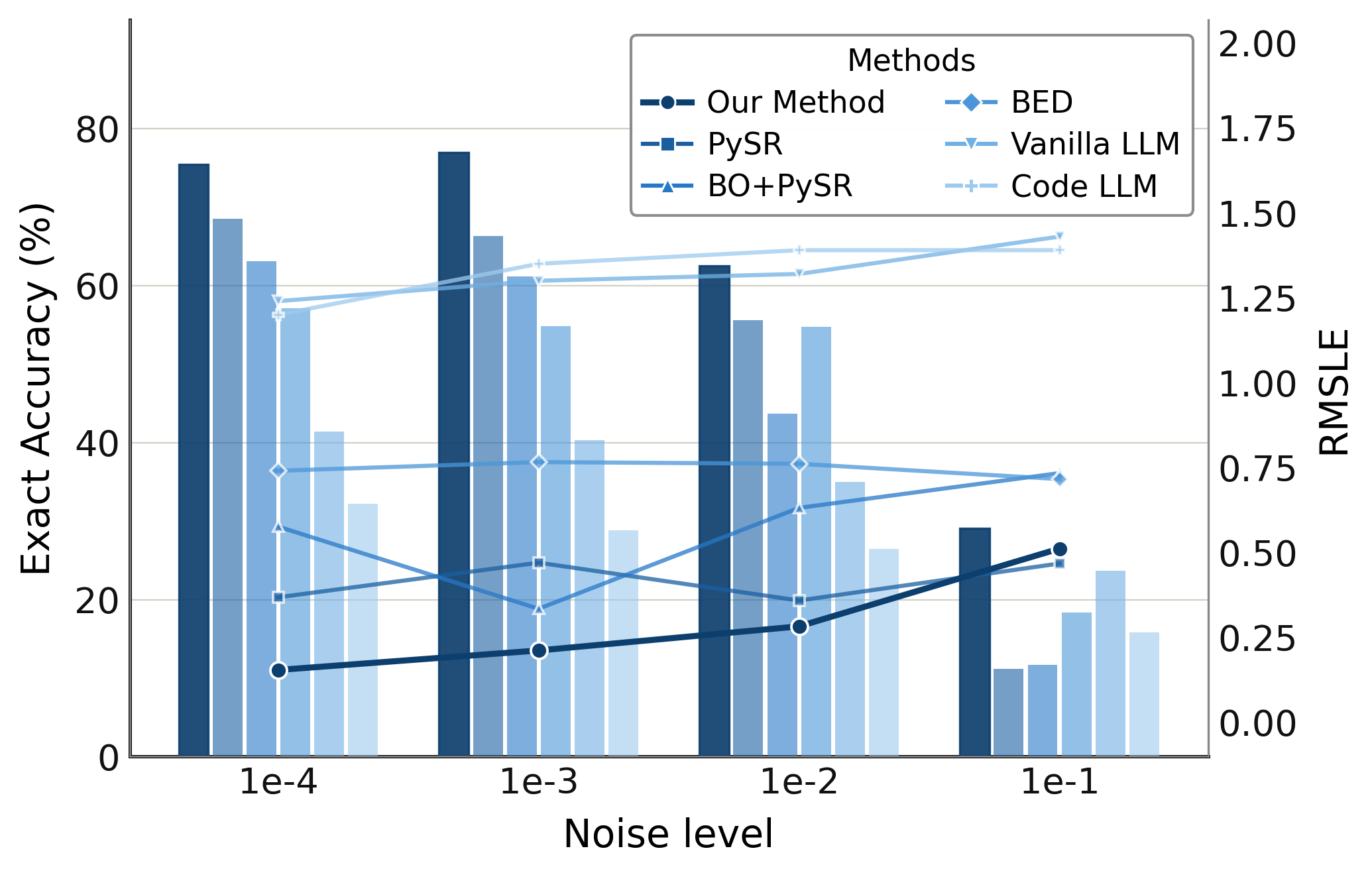}
        \caption{NewtonBench}
        \label{fig:noise_newton}
    \end{subfigure}
    \hfill
    \begin{subfigure}[t]{0.32\linewidth}
        \centering
        \includegraphics[width=\linewidth]{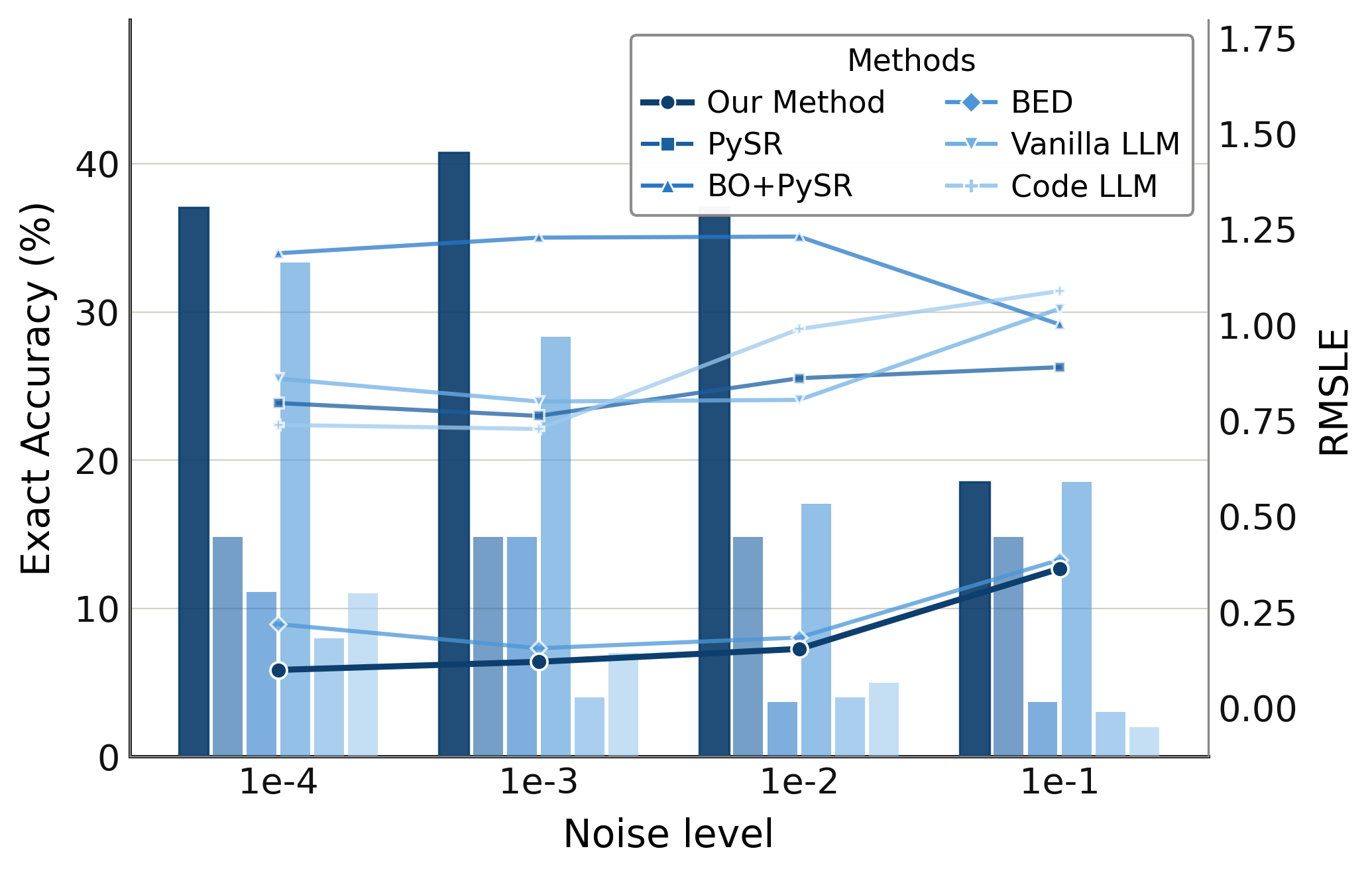}
        \caption{\chem}
        \label{fig:noise_chem}
    \end{subfigure}
    \hfill
    \begin{subfigure}[t]{0.32\linewidth}
        \centering
        \includegraphics[width=\linewidth]{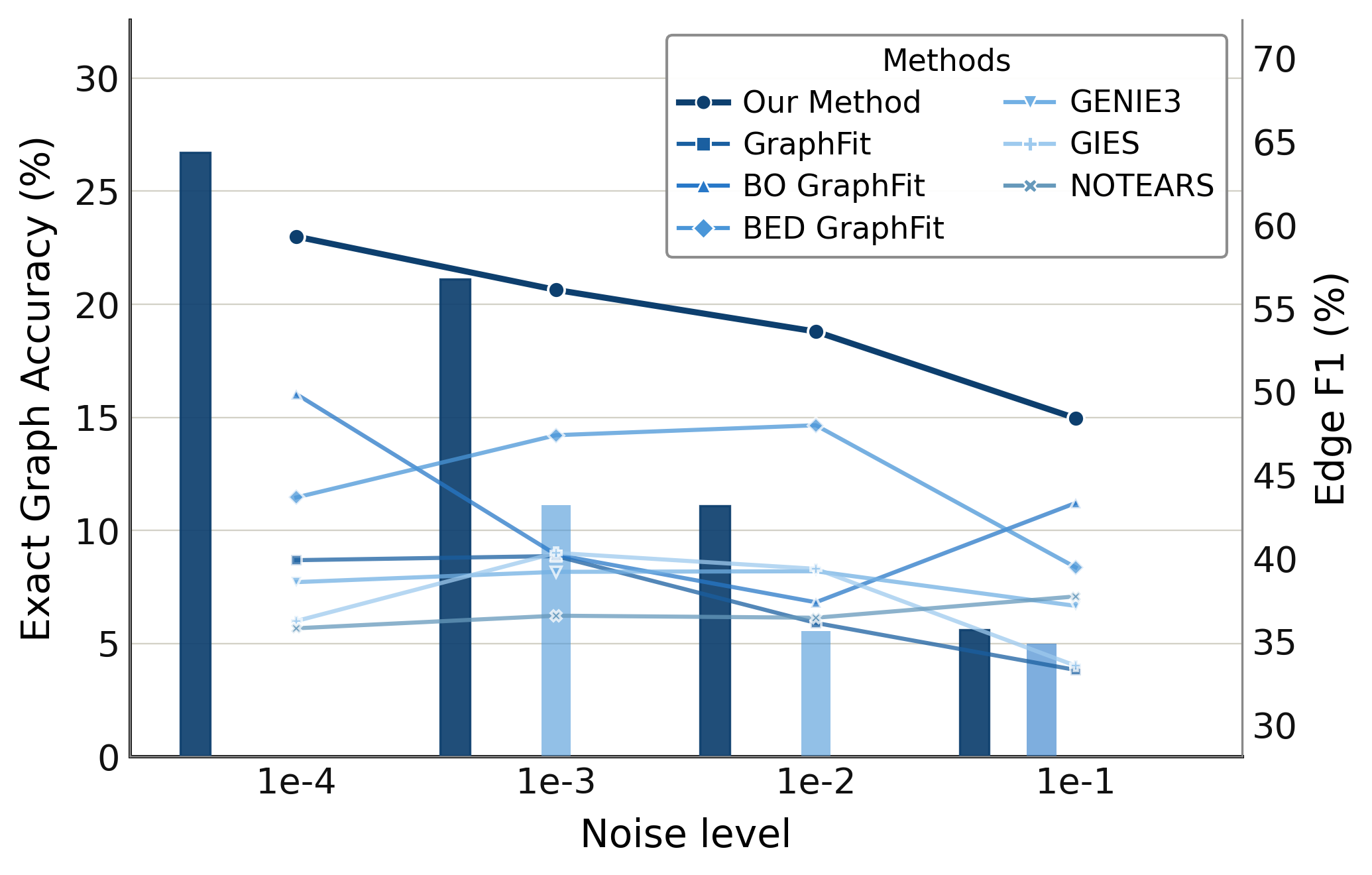}
        \caption{\grn}
        \label{fig:noise_grn}
    \end{subfigure}

    \caption{
    Robustness to observation noise across NewtonBench, \chem, and \grn.
    Bars report exact accuracy, while lines report the continuous error or graph recovery metric:
    RMSLE for NewtonBench and \chem, and Edge F1 for \grn.
    }
    \label{fig:noise_robustness}
\end{figure}

Figure \ref{fig:noise_robustness} reports a noise sensitivity analysis evaluating robustness to increasing levels of observation noise across all three benchmarks. For each benchmark, we inject controlled Gaussian noise at varying levels and measure the effect on both primary recovery metrics and predictive fidelity. Across all settings, \modelname shows stronger recovery than baselines as noise increases, consistent with scientific priors that constrain the search to physically plausible structures even as data quality degrades. On NewtonBench and \chem, symbolic accuracy exhibits threshold behavior, degrading in a step-function pattern while RMSLE increases smoothly, reflecting a transition between a regime where noise blurs parameter estimates but structural identifiability is preserved, and a regime where competing hypotheses become observationally indistinguishable within the current budget. On \grn, the two recovery metrics decouple sharply under noise. Exact graph accuracy drops severely even at moderate noise levels, while edge F1 deteriorates slowly and gradually. This dissociation reveals that noise primarily disrupts precise topology recovery while the broader edge-level structure remains partially identifiable. The result suggests that activation versus repression provides a stronger, more noise-resilient signal than exact graph structure and points to precise topology recovery as the primary fragility of the current framework under noisy perturbation settings.

\subsection{Model Capability}
\label{app:model_capability}

\begin{wraptable}{r}{0.50\linewidth}
\vspace{-1.0em}
\centering
\captionsetup{font=small}
\caption{\textbf{LLM backbone comparison.} GPT-4o-mini is the default backbone; Qwen3 models evaluate open-weight scaling.}
\label{tab:qwen_capability}
\setlength{\tabcolsep}{2.5pt}
\renewcommand{\arraystretch}{1.05}
\tiny
\resizebox{0.95\linewidth}{!}{%
\begin{tabular}{l ccc ccc ccc}
\toprule
& \multicolumn{3}{c}{\textbf{Newton}}
& \multicolumn{3}{c}{\textbf{Chem}}
& \multicolumn{3}{c}{\textbf{GRN}} \\
\cmidrule(lr){2-4} \cmidrule(lr){5-7} \cmidrule(lr){8-10}
\textbf{Backbone}
& \textbf{SA} & \textbf{Ex.} & \textbf{RMSLE}
& \textbf{SA} & \textbf{Ex.} & \textbf{RMSLE}
& \textbf{F1} & \textbf{Ex.} & \textbf{Sign} \\
\midrule
GPT-4o-mini
& \textbf{67.60} & 81.50 & 0.150
& \textbf{35.09} & 50.88 & 0.189
& \textbf{72.49} & \textbf{31.11} & 98.15 \\

\midrule
Qwen3-4B
& 57.78 & 82.86 & 0.174
& 12.92 & 53.56 & 0.3766
& 49.56 & 21.78 & 93.52 \\

Qwen3-14B
& 54.29 & 86.11 & 0.127
& 23.88 & 50.59 & 0.1574
& 51.19 & 22.34 & 92.16 \\

Qwen3-32B
& 60.56 & \textbf{88.12} & \textbf{0.101}
& 25.56 & \textbf{52.81} & \textbf{0.0815}
& 62.70 & 25.88 & \textbf{98.70} \\
\bottomrule
\end{tabular}%
}
\vspace{-1.2em}
\end{wraptable}

Table~\ref{tab:qwen_capability} evaluates the effect of LLM backbone capability on closed-loop discovery. We keep the \texttt{Qwen2.5-7B-Instruct} as the smaller local ensemble model, but vary the primary model driving the experiments. Model scale is most beneficial on the more compositional and structured benchmarks. \chem improves consistently from \texttt{Qwen3-4B} to \texttt{Qwen3-32B} across SA, exact accuracy, and RMSLE, while \grn shows clear gains in edge F1 and exact graph accuracy. This suggests that stronger backbones provide better mechanistic priors for selecting relevant variables, proposing discriminative experiments, and reasoning over structured mechanisms.
NewtonBench reveals a different pattern. \texttt{Qwen3-32B} achieves the lowest RMSLE and highest exact accuracy, but \texttt{Qwen3-14B} obtains higher symbolic accuracy. Thus, larger models can fit behavior more accurately without always recovering the exact symbolic form. Since the difference is small, this should not be interpreted as evidence against larger models; rather, the non-monotonic symbolic trend suggests that closed-loop discovery is not determined by scale alone, but by the interaction between hypothesis generation, experiment selection, and refinement.

\subsection{Relative Sample Efficiency}
\label{relative_query_cost}

\begin{figure*}[!htbp]
    \centering
    \includegraphics[width=\textwidth]{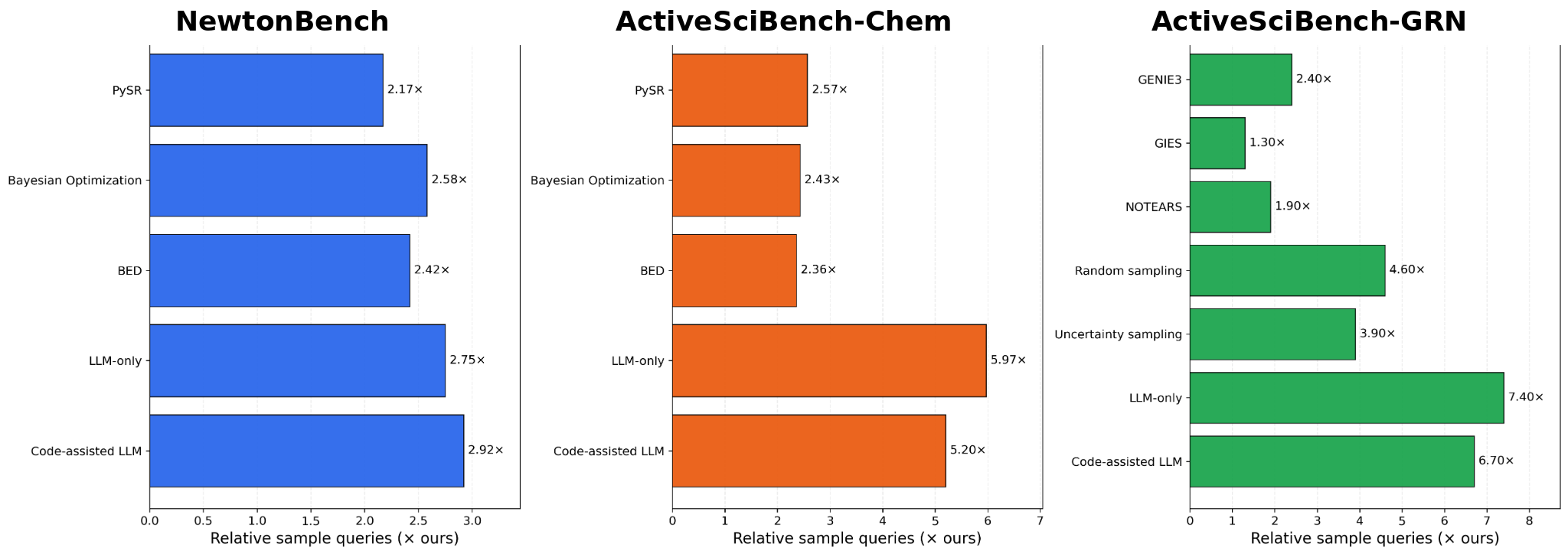}
    \caption{\small \textbf{Relative sample efficiency across benchmarks.} Each bar shows the multiplicative number of samples required by a comparison method to match the fixed-budget performance of \modelname{}~(lower is better). The number of samples is measured relative to the reference budgets used for \modelname{}: \(B{=}20\) for NewtonBench, \(B{=}60\) for \chem, and \(B{=}20\) for \grn.}
    \label{fig:relative_query_cost}
\end{figure*}

Figure~\ref{fig:relative_query_cost} reports target-matching relative sample efficiency across the three benchmark families. For each benchmark, we treat the fixed-budget performance of \modelname{} as the reference target and measure how many oracle queries a comparison method requires to match that target. A relative sample efficiency of \(2\times\) therefore means that the comparison method requires twice as many queries as \modelname{} to attain the same level of performance.

On NewtonBench, all comparison methods require more than twice the query budget of \modelname{}, with symbolic regression baselines ranging from \(2.17\times\) to \(2.58\times\) and LLM-only variants ranging from \(2.75\times\) to \(2.92\times\). On \chem, the strongest non-LLM methods remain above \(2.3\times\), while the LLM-only and code-assisted LLM conditions require \(5.97\times\) and \(5.20\times\) the reference budget, respectively. On \grn, the canonical graph baselines GENIE3, GIES, and NOTEARS require \(2.4\times\), \(1.3\times\), and \(1.9\times\) the reference budget, while active and LLM-based alternatives require between \(3.9\times\) and \(7.4\times\). Taken together, these results show that the main benefit of \modelname{} is not only improved final recovery but also substantially better budget utilization, since conditioning acquisition on explicitly competing mechanistic explanations makes each oracle query more informative for resolving structural ambiguity.

\subsection{Qualitative Analysis}
\begin{figure}[!htbp]
    \centering
    \includegraphics[width=0.98\textwidth]{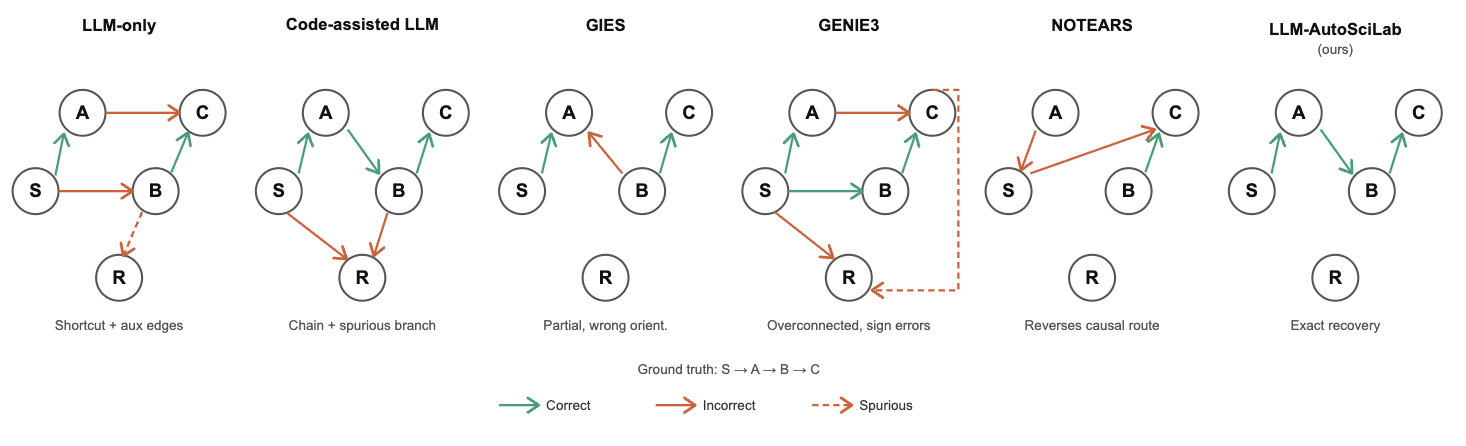}
    \caption{\small \textbf{Qualitative \grn case study.}
    \modelname exactly recovers the sparse activation chain, while baselines either add spurious auxiliary edges, reverse edge orientation, or recover only partial structure. Green edges indicate correctly recovered relations; red edges indicate incorrect or spurious relations.}
    \label{fig:qual_grn}
\end{figure}

Figure~\ref{fig:qual_grn} shows a representative \grn example where the target mechanism is a sparse activation chain, $S \rightarrow A \rightarrow B \rightarrow C$, with an irrelevant regulator $R$. \modelname exactly recovers the ground-truth chain and correctly excludes $R$, indicating that its perturbation choices isolate the causal backbone rather than merely fitting correlated responses. In contrast, the baselines recover only parts of the structure. LLM-only adds shortcut and auxiliary edges, code-assisted LLM preserves the main chain but introduces a spurious side branch, and classical graph-discovery methods either overconnect the graph, reverse orientations, or miss key dependencies. This example illustrates the main GRN failure mode: baselines often detect local associations but fail to distinguish direct causal edges from indirect or irrelevant perturbation effects, whereas hypothesis-conditioned acquisition supports targeted disambiguation of the signed graph structure.

\subsection{Statistical significance}
\label{app:stats}
\begin{table*}[!htbp]
\centering
\fontsize{8}{8}\selectfont
\caption{\small \textbf{Standard Deviation across benchmarks.} Values are reported as mean $\pm$ standard deviation across 3 seeds. For NewtonBench and \chem, columns are SA / Exact / RMSLE. For \grn, columns are F1 / Exact / Sign.}
\label{tab:across_seed_variability}
\renewcommand{\arraystretch}{1}
\setlength{\tabcolsep}{4pt}
\begin{tabular}{lccc}
\toprule
\textbf{Method} & \textbf{SA} & \textbf{Ex.} & \textbf{RMSLE} \\
\midrule
\multicolumn{4}{c}{\textit{\textbf{NewtonBench}}} \\
PySR & 24.07 $\pm$ 0.00 & 74.54 $\pm$ 1.96 & 0.182 $\pm$ 0.008 \\
BO+PySR & 24.54 $\pm$ 0.65 & 68.52 $\pm$ 2.62 & 0.334 $\pm$ 0.023 \\
BED & 11.11 $\pm$ 0.00 & 63.86 $\pm$ 1.09 & 0.577 $\pm$ 0.044 \\
LLM-only & 6.48 $\pm$ 0.40 & 7.41 $\pm$ 0.80 & 5.039 $\pm$ 0.156 \\
Code-assisted LLM & 7.41 $\pm$ 0.50 & 10.19 $\pm$ 0.20 & 4.912 $\pm$ 0.128 \\
\rowcolor{gray!10}
\textbf{LLM-AutoSciLab (Ours)} & \textbf{67.60 $\pm$ 0.20} & \textbf{81.50 $\pm$ 1.30} & \textbf{0.150 $\pm$ 0.003} \\
\midrule
\multicolumn{4}{c}{\textit{\textbf{\chem}}} \\
PySR & 7.89 $\pm$ 0.00 & 15.79 $\pm$ 1.20 & 1.212 $\pm$ 0.016 \\
Bayesian Optimization & 5.26 $\pm$ 0.40 & 6.98 $\pm$ 2.60 & 1.960 $\pm$ 0.288 \\
BED & 31.58 $\pm$ 0.00 & 39.47 $\pm$ 1.50 & 0.249 $\pm$ 0.015 \\
LLM-only & 0.00 $\pm$ 0.00 & 0.00 $\pm$ 0.00 & 0.764 $\pm$ 0.189 \\
Code-assisted LLM & 0.00 $\pm$ 0.00 & 0.00 $\pm$ 0.00 & 0.824 $\pm$ 0.183 \\
\rowcolor{gray!10}
\textbf{LLM-AutoSciLab (Ours)} & \textbf{35.09 $\pm$ 0.80} & \textbf{50.88 $\pm$ 2.60} & \textbf{0.189 $\pm$ 0.004} \\
\midrule
\textbf{Method} & \textbf{Edge F1} & \textbf{Exact Graph} & \textbf{Sign} \\
\midrule
\multicolumn{4}{c}{\textit{\textbf{\grn}}} \\
GENIE3 & 38.60 $\pm$ 0.00 & 0.00 $\pm$ 0.00 & 79.07 $\pm$ 1.60 \\
GIES & 56.27 $\pm$ 1.80 & 6.67 $\pm$ 1.33 & 85.78 $\pm$ 2.30 \\
NOTEARS & 27.60 $\pm$ 1.20 & 2.22 $\pm$ 0.00 & 58.15 $\pm$ 1.30 \\
Random sampling & 35.56 $\pm$ 3.80 & 2.22 $\pm$ 0.00 & 84.44 $\pm$ 2.00 \\
Uncertainty sampling & 50.10 $\pm$ 1.40 & 4.44 $\pm$ 0.00 & 88.15 $\pm$ 2.00 \\
LLM-only & 50.41 $\pm$ 3.50 & 0.00 $\pm$ 0.00 & 79.44 $\pm$ 0.60 \\
Code-assisted LLM & 54.67 $\pm$ 4.70 & 0.00 $\pm$ 0.00 & 85.98 $\pm$ 0.90 \\
\rowcolor{gray!10}
\textbf{LLM-AutoSciLab (Ours)} & \textbf{72.49 $\pm$ 1.30} & \textbf{31.11 $\pm$ 1.33} & \textbf{98.15 $\pm$ 0.30} \\
\bottomrule
\end{tabular}
\end{table*}

To assess statistical significance, we repeated each benchmark configuration across three seeds and report the mean and standard deviation of the seed-level aggregate scores. The mean results are presented in Table~\ref{tab:main_results}. For NewtonBench and \chem, we report symbolic accuracy (SA), exact accuracy, and RMSLE. For \grn, we report edge $F_1$, exact graph accuracy, and sign accuracy. Overall, the variance across benchmarks is modest. The strongest methods, including \modelname{}, remain stable across seeds, while weaker LLM-only or unguided baselines show somewhat larger variability, particularly on harder graph-recovery settings. Standard deviations of $\pm 0.00$ indicate that the seed-level aggregate scores are identical at the displayed precision as expected for deterministic baselines or methods run under fixed random seeds.

\section{Benchmark Details}
\label{appendix_benchmark_details}

\subsection{\chem}
\label{appendix_chem}

\chem is an active enzyme kinetics benchmark for mechanism recovery under finite experimental budgets. Rather than receiving a fixed dataset, the learner adaptively selects biochemical assay conditions and observes reaction rates to infer hidden kinetic mechanisms and parameters. This setup reflects the challenge of scientific discovery, where multiple mechanisms can explain limited observations and targeted experiments are required to distinguish competing hypotheses.
\[
\mathbf{x}=(C_A, C_I, C_B, C_P, \mathrm{Enz}, T, \mathrm{pH}),
\]
covering substrate, inhibitor, secondary substrate, product concentration, enzyme loading, temperature, and pH. While every task exposes the same interface, each mechanism depends on only a subset of variables, requiring the learner to identify both the governing law and the relevant dimensions. Dependence on different variables corresponds to distinct mechanistic behaviors such as inhibition, bisubstrate reactions, product feedback, or environmental modulation. \chem contains \textbf{57 curated tasks} organized into easy, medium, and hard tiers. Easy tasks correspond to standard kinetic families, medium tasks introduce structured compositions, and hard tasks include weaker identifiability and nonstandard behaviors such as cooperative or allosteric kinetics. 
The \chem benchmark evaluates both active experimentation and symbolic scientific reasoning and is organized around \textbf{nine canonical base families}:
\begin{itemize}[leftmargin=*]
\setlength\itemsep{-0.2em}
\item{Michaelis--Menten saturation}
\item{Competitive inhibition}
\item{Product inhibition}
\item{Arrhenius temperature dependence}
\item{Ping-pong bisubstrate kinetics}
\item{Uncompetitive inhibition}
\item{Substrate inhibition}
\item{Hill cooperativity}
\item{Noncompetitive inhibition}
\end{itemize}
Beyond the nine base families, \chem includes \textbf{structured composite mechanisms} that combine substrate-response kinetics with modifiers such as inhibition, temperature dependence, or feedback. Examples include Michaelis--Menten with competitive inhibition and Arrhenius modulation, ping-pong bisubstrate kinetics with noncompetitive inhibition, and Hill cooperativity with product feedback. These composites move beyond isolated textbook mechanisms and require the learner to distinguish competing mechanistic explanations under a shared assay interface. \chem also includes a targeted set of \textbf{extended and nonstandard mechanism families} beyond the core base library:

\begin{itemize}[leftmargin=*]
\setlength\itemsep{-0.2em}
\item{Ordered sequential bisubstrate kinetics}
\item{Allosteric activation}
\item{Anti-cooperative Hill behavior}
\item{Fractal or anomalous kinetics}
\item{Mixed inhibition}
\item{Cooperative inhibition}
\item{Monotonic pH dependence}
\item{Metal-ion activation}
\item{Product activation / autocatalytic feedback}
\item{Dual inhibition by inhibitor and product}
\end{itemize}

Taken together, the 57 \chem tasks span textbook kinetic families, structured compositions, and targeted extended mechanisms beyond the standard library. This makes \chem both interpretable at the mechanism level and sufficiently rich to require active experimentation, relevant-variable identification, and mechanistic discrimination rather than simple equation retrieval. \chem is a simulator-based oracle benchmark designed to isolate active kinetic mechanism discovery under controlled, reproducible conditions. Observations are generated from mechanistically specified kinetic families with hidden parameters and benchmark-defined noise, rather than physical wet-lab experiments. While it does not capture the full complexity of laboratory biochemistry, such as assay failures, batch effects, protocol variability, or experimental cost, it provides a clean, budget-controlled setting for evaluating closed-loop mechanistic recovery.


\subsection{\grn}
\label{appendix_grn}

\grn is an online gene perturbation benchmark for active causal graph discovery in gene regulation. Unlike \chem, which focuses on recovering biochemical rate laws, \grn requires the learner to infer hidden causal structure and nonlinear dynamics from interventional data. Each task simulates a hidden regulatory system with unknown graph structure, edge signs, and nonlinear dynamics. The learner performs discrete interventions, such as gene knock-up or knock-down experiments, and observes downstream expression changes, while the underlying graph remains unobserved. Different motifs can produce similar responses under limited interventions, and hidden parameters can make identical motifs appear qualitatively different, requiring the learner to jointly infer topology, sign, and dynamics from sparse experimental data. \grn is built from canonical regulatory motifs spanning increasing structural complexity, including activation, repression, feedback, and switching behavior. The paper-facing benchmark contains five core motif families, three topological variants, and three difficulty levels, yielding 45 tasks per random seed. These motifs are standard systems biology primitives that provide a compact yet meaningful testbed for mechanistic graph discovery.
The five \grn motif families are:
\setlength\itemsep{-0.2em}
\begin{itemize}[leftmargin=*]
\item{\textbf{Activation chain.}} A layered activation cascade from signal to intermediate regulators to the reporter.
\item{\textbf{Coherent feedforward loop.}} A motif in which the input acts through both a direct and a mediated activating branch.
\item{\textbf{Incoherent feedforward loop.}} A motif in which activation and repression act along competing paths, producing adaptation-like or pulse-like behavior.
\item{\textbf{Negative-feedback circuit.}} A self-limiting repression architecture in which downstream activation induces a repressive branch.
\item{\textbf{Toggle-switch or bistable decision circuit.}} A mutually repressive switching architecture used to model bistability, state selection, and cellular decision making.
\end{itemize}

Each motif family is instantiated across three topological variants and three difficulty levels. The topological variants preserve the motif identity while changing the precise wiring structure, whereas the difficulty levels correspond to increasingly nonlinear and feedback-sensitive parameter regimes. Easy tasks exhibit near-linear responses that reveal the graph relatively clearly, while medium and hard tasks introduce saturation, switching behavior, and bistability that obscure weak or mediated dependencies. As a result, \grn is not simply a passive graph estimation problem, but a budgeted intervention design task in which the learner must select perturbations that best distinguish competing regulatory mechanisms. \grn is a simulator-based oracle benchmark designed to isolate intervention-driven regulatory discovery under controlled, reproducible conditions. Responses are generated from motif-specific nonlinear dynamics with hidden parameters and benchmark-defined noise rather than physical biological experiments. Although it does not capture the full complexity of real perturbation biology, including failed interventions, cell-state heterogeneity, batch effects, off-target effects, or experimental cost variability, it provides a clean, budget-controlled testbed for evaluating closed-loop mechanistic graph recovery.

\vspace{-0.5em}
\subsection{NewtonBench}
\vspace{-0.5em}
NewtonBench provides the physics component of our benchmark suite. In contrast to \chem and \grn, it isolates active symbolic law discovery in a setting where the relevant variables for each task are already known. In our experiments, one oracle call evaluates the hidden physical law at a chosen assignment of task-specific input variables, and the learner must recover the symbolic law under a finite query budget. Because NewtonBench is already introduced as a standalone benchmark in \cite{zheng2026newtonbench}, we refer readers there for the full benchmark construction, counterfactual law-generation procedure, and task catalog. Table \ref{tab:benchmark_summary_long} summarizes the benchmark suite we use for evaluation in this work.
\begin{table*}[!htbp]
\centering
\small
\caption{\small \textbf{Benchmark summary.} NewtonBench, \chem, and \grn differ in scientific setting, discovery target, interface, and benchmark families.}
\label{tab:benchmark_summary_long}
\renewcommand{\arraystretch}{1.15}
\setlength{\tabcolsep}{6pt}
\begin{tabularx}{\textwidth}{p{2.7cm} X}
\toprule
\textbf{Attribute} & \textbf{Description} \\
\midrule

\multicolumn{2}{l}{\textbf{NewtonBench}} \\
\midrule
Scientific setting & Active physics law discovery in known-variable domains. \\
Discovery target & Symbolic physical law. \\
Interface & Task-specific physical variables; one oracle call evaluates the hidden law at a chosen assignment of inputs. \\
Setting used in this work & \texttt{vanilla\_equation}. \\
Core domains & Newtonian gravitation; Coulomb force; magnetic force; Fourier heat conduction; Snell's law; radioactive decay; underdamped harmonic motion; Malus's law; speed of sound; Hooke's law; Bose--Einstein distribution; heat transfer. \\
Domain count & 12. \\
Role in benchmark suite & Isolates active symbolic law discovery when the relevant variables are already known. \\
\addlinespace[0.4em]
\midrule
\multicolumn{2}{l}{\textbf{\chem}} \\
\midrule
Scientific setting & Active enzyme-kinetics discovery under a shared biochemical assay interface. \\
Discovery target & Symbolic kinetic law with hidden relevant-variable structure. \\
Interface & Shared 7D biochemical assay interface:
$(C_A, C_I, C_B, C_P, \mathrm{Enz}, T, \mathrm{pH})$. \\
Primary observation & Initial rate $r_0$ together with auxiliary mass-balance observables. \\
Base families & Michaelis--Menten saturation; competitive inhibition; product inhibition; Arrhenius temperature dependence; ping-pong bisubstrate kinetics; uncompetitive inhibition; substrate inhibition; Hill cooperativity; noncompetitive inhibition. \\
Composite mechanisms & Structured compositions of the base families, combining substrate-response families with modifiers such as inhibition, temperature dependence, feedback, and bisubstrate effects. Representative examples include MM with competitive inhibition and Arrhenius modulation, ping-pong with noncompetitive inhibition, Hill cooperativity with product modulation, and substrate inhibition with Arrhenius modulation. \\
Extended families & Ordered sequential bisubstrate kinetics; allosteric activation; anti-cooperative Hill behavior; fractal / anomalous kinetics; mixed inhibition; cooperative inhibition; monotonic pH dependence; metal-ion activation; product activation / autocatalytic feedback; dual inhibition by inhibitor and product. \\
Domain count & 57 curated domain classes in the paper-facing benchmark. \\
Role in benchmark suite & Adds hidden relevant-variable discovery and mechanistically grounded biochemical ambiguity beyond known-variable symbolic regression. \\
\addlinespace[0.4em]
\midrule
\multicolumn{2}{l}{\textbf{\grn}} \\
\midrule
Scientific setting & Intervention-driven regulatory discovery from perturbation-response experiments. \\
Discovery target & Signed regulatory graph / motif together with effective nonlinear regulatory structure. \\
Interface & Shared perturbation interface with discrete interventions over hidden regulatory systems. \\
Primary observation & Downstream expression changes following intervention. \\
Motif families & Activation chain; coherent feedforward loop; incoherent feedforward loop; negative-feedback circuit; toggle-switch / bistable decision circuit. \\
Topological variants & Three versions per motif family. \\
Difficulty levels & Easy, medium, and hard parameter regimes, corresponding to progressively stronger nonlinearity and more difficult motif disambiguation. \\
Family count & 5 core motif families. \\
Role in benchmark suite & Adds intervention-driven signed graph discovery and causal-mechanism ambiguity beyond equation recovery alone. \\
\bottomrule
\end{tabularx}
\end{table*}

\vspace{-0.5em}

\section{Metrics}
\label{appendix_metrics}
\vspace{-0.5em}
We evaluate all three benchmarks under a unified objective of recovering the true scientific mechanism. For NewtonBench and \chem, we report three metrics. First, we measure predictive fidelity using the root mean squared logarithmic error (RMSLE),
\vspace{-0.25em}
\[
\mathrm{RMSLE}(\hat{f}, f)
=
\sqrt{\frac{1}{N}\sum_{i=1}^{N}
\left(\log(1+\hat{y}_i)-\log(1+y_i)\right)^2 }.
\]
\vspace{-0.25em}
\hspace{-0.25em}RMSLE is appropriate because target values in these domains can span multiple orders of magnitude. Second, we report numerical exact accuracy,
\vspace{-0.15em}
\[
\mathrm{ExAcc}(\hat{f})
=
\mathbf{1}\!\left[\mathrm{RMSLE}(\hat{f}, f) < 0.01\right],
\]
\vspace{-0.15em}
\hspace{-0.25em}which follows the paper's exact-recovery criterion. Third, we report symbolic accuracy (SA), which measures whether the recovered law matches the ground-truth mechanism up to algebraic rewriting, fitted constants, and variable renaming. Symbolic accuracy is stricter than numerical exactness, since a numerically accurate approximation need not recover the correct mechanistic form. For \grn, the target is graph recovery rather than scalar law recovery. We therefore report edge-level precision, recall, and F1, sign accuracy for activation versus repression, exact graph accuracy, and motif accuracy. These metrics distinguish partial structural recovery from full mechanistic recovery. For NewtonBench and \chem, symbolic accuracy is evaluated with an LLM judge that determines whether the predicted hypothesis is equivalent to the ground-truth expression up to constant parameter values. In this evaluation, the ground-truth law is presented as expression $A$, and the candidate hypothesis $B$ may be represented either as an executable program or as a symbolic expression. As illustrated, the judge is prompted as follows:
\vspace{-0.5em}
\begin{verbatim}
Question: Given the ground truth mathematical expression A and the 
hypothesis B, determine if there exist any constant parameter values that 
would make the hypothesis equivalent to the given ground truth expression. 
Let's think step by step. Explain your reasoning and 
then provide the final answer as:
{
  "reasoning": "Brief step-by-step analysis",
  "answer": "Yes/No"
}
\end{verbatim}

A prediction is counted as symbolically correct if the judge returns ``Yes.'' This metric is stricter than numerical exactness, since a numerically accurate approximation need not recover the same underlying symbolic form.

\section{\benchmark Design and Memorization}
\label{memorization_appendix}
A key concern for LLM-based scientific discovery is whether strong performance reflects genuine mechanistic recovery or simple recall of textbook systems. Our benchmarks are designed to make retrieval alone insufficient. Each run begins without observations, requiring the learner to actively select experiments under a limited oracle budget and infer the hidden mechanism from the resulting responses. The instantiated system, not just the family label, is hidden throughout. In NewtonBench, the learner observes the natural variables but not the hidden governing law. In \chem, all tasks share a common seven-variable assay interface while the relevant variables, mechanism class, and parameterization remain hidden. In \grn, the learner must infer a hidden signed regulatory mechanism from perturbation responses rather than directly observing the graph. At the same time, realistic scientific semantics are preserved through variables such as substrate, inhibitor, temperature, and perturbation target, ensuring the evaluation measures scientific reasoning rather than anonymized symbol matching. This can also be quantified through conservative lower bounds on the hidden task space. In \chem, the 57 mechanism classes contain between 2 and 8 hidden continuous parameters (median 5). Even under a coarse discretization of only 100 possible values per parameter, this yields more than $6 \times 10^{16}$ possible hidden instantiations. \grn is broader still: the target is a signed regulatory graph over ${signal, A, B, C, R}$ with 20 possible directed non-self edges, each absent, activating, or repressing, yielding $3^{20} \approx 3.5 \times 10^9$ unconstrained signed graphs. Even with a sparsity cap of six edges, this leaves more than 3 million candidate graphs before accounting for hidden continuous dynamics. These values are conservative lower bounds, but they illustrate that the benchmarks cannot be reduced to retrieving a small library of familiar mechanisms.

\section{Implementation Details}
\subsection{\modelname{}}
\label{appendix_autoscilab_method}
\begin{wrapfigure}{r}{0.48\textwidth}
\vspace{-2.50em}
\begin{minipage}{\linewidth}
\begin{algorithm}[H]
\small
\caption{Adaptive Ensemble}
\label{alg:adaptive_ensemble}
\begin{algorithmic}[1]
\Require Prompt context $S_t$, base ensemble size $K$, cap $K_{\max}$, entropy threshold $\tau_H$
\State $\mathcal{C} \gets \emptyset,\; H_{\mathrm{prev}} \gets \varnothing$
\While{$|\mathcal{C}| < K_{\max}$}
    \State $b \gets \min(K, K_{\max}-|\mathcal{C}|)$
    \State $\mathcal{B} \gets \texttt{SampleHypotheses}(\pi_\theta^{\rm small}, S_t, b)$
    \State $\mathcal{C} \gets \mathcal{C} \cup \texttt{FilterValid}(\mathcal{B})$
    \If{$|\mathcal{C}| < K$}
        \State \textbf{continue}
    \EndIf
    \Statex \small \textcolor{gray}{{\# {Cluster $\mathcal{C}$ by structural skeleton and compute entropy $H$}}}
    \If{$H_{\mathrm{prev}} \neq \varnothing$}
        \If{$|H-H_{\mathrm{prev}}| < \tau_H$}
            \State \textbf{break}
        \EndIf
    \EndIf
    \State $H_{\mathrm{prev}} \gets H$
\EndWhile
\State \Return $\texttt{BuildDistribution}(\mathcal{C})$
\end{algorithmic}
\end{algorithm}
\end{minipage}
\vspace{-2.5em}
\end{wrapfigure}
Our implementation of \modelname{} instantiates the closed-loop discovery framework in a hypothesis-driven setting with adaptive ensembling. At each stage, the system maintains a set of candidate mechanisms, proposes informative experiments conditioned on disagreement among those candidates, updates the observation set with oracle feedback, and periodically refines the candidate laws through symbolic fitting. The implementation uses a two-model architecture in which a smaller local model generates a diverse hypothesis set, while the main model synthesizes these hypotheses with accumulated evidence to guide subsequent experimentation. Symbolic refinement results are summarized in a structured memory representation and re-injected into later reasoning steps.

\paragraph{Prompting and Hypothesis Generation.}

The hypothesis-generation step receives the current goal, domain description, parameter names, Python function signature, experiment history as a text table, the current working hypothesis, the remaining budget, the current phase, the best symbolic equation found so far (if available), and the structured memory summary. At this stage, the model is asked to propose one primary hypothesis together with multiple alternate hypotheses, but \emph{not} search regions. All hypotheses are returned as pure Python expressions using the exact oracle variable names and symbolic free constants such as \texttt{C0}, \texttt{C1}, \texttt{alpha}, and \texttt{beta}. In the paper configuration, the smaller ensemble model is sampled in parallel with base ensemble size $K=5$, and each call produces a structured hypothesis output from which the primary candidate is retained for ensemble construction.

\paragraph{Adaptive Ensemble Construction.}

Raw ensemble hypotheses are filtered before use: expressions that cannot be executed on the observed data, produce predominantly non-finite predictions, or are clearly nonsensical under the current dataset are discarded. The remaining hypotheses are then clustered by structural skeleton, obtained by canonicalizing constants while preserving functional form. \modelname{} uses adaptive ensemble growth in which hypotheses are sampled in batches of size $K$, reclustered after each batch, and the Shannon entropy of the structural cluster distribution is recomputed. Sampling stops when the entropy change between successive batches falls below $0.1$, or when the hard cap of $K_{\max}=20$ samples is reached. The resulting hypothesis distribution stores the raw hypotheses, structurally unique representatives, cluster assignments, the majority-cluster agreement score, and a synthesis summary that is passed to the main model.

\paragraph{Hypothesis-Conditioned Acquisition.}

The main model receives a second prompt containing the full experiment history, the current working hypothesis, the structured memory summary, and a compact summary of the current hypothesis set, including the number of sampled hypotheses, the number of unique structures, and representative candidates from each structural cluster. It returns an updated primary hypothesis, alternate hypotheses, and a set of search regions. In parallel, \modelname{} computes a falsification-oriented disagreement score directly from the current hypothesis set. For a candidate point $\mathbf{x}$, let $\hat{f}_1(\mathbf{x}),\dots,\hat{f}_K(\mathbf{x})$ denote the predictions induced by the current candidate mechanisms. We define the disagreement score as
\[
\Delta(\mathbf{x})
=
\mathrm{Std}\!\left(
\log_{10}\hat{f}_1(\mathbf{x}),
\ldots,
\log_{10}\hat{f}_K(\mathbf{x})
\right).
\]
\noindent
For \grn, disagreement is computed over fitted graph hypotheses through
their predicted intervention responses. If graph hypothesis $g_k$ predicts a
post-intervention response vector $\hat{\by}_k(a)\in\mathbb{R}_{>0}^d$ for
intervention $a$, we define
\[
\delta_{\mathrm{GRN}}(a)
=
\frac{1}{d}\sum_{\ell=1}^{d}
\operatorname{Var}_{k=1,\ldots,K}\!\Bigl[\log\!\bigl(\hat{y}_{k,\ell}(a)+\varepsilon\bigr)\Bigr].
\]
This means that edge and sign disagreements matter only through their
falsifiable intervention consequences. The role of the LLM at this stage is to propose search regions expected to be informative for separating competing mechanisms; the final experiment points are then selected by the active-learning layer within those regions. When the system is in the low-confidence regime, candidate points are sampled from the proposed bounds, scored by $\Delta(\mathbf{x})$, and a small diverse subset is chosen for oracle evaluation. This yields a hypothesis-conditioned acquisition rule that preferentially queries regions where competing mechanisms make sharply different predictions.

\paragraph{Hypothesis Refinement.}
Mechanism refinement is performed on accumulated observations using a domain-specific refinement backend. The refinement stage is run periodically rather than continuously. Before the backend is invoked, the loop extracts the variables implicated by the current hypothesis set and uses them to focus the refinement search, optionally reintroducing variables whose residual behavior suggests missing structure. In equation-discovery settings, refinement combines direct fitting of candidate structural families with numerical parameter optimization and symbolic search over the accumulated observations. In graph-discovery settings, refinement updates the candidate signed regulatory structures and their associated dynamical parameters using the observed perturbation responses. The refined candidates are then pooled for later arbitration and memory updates.

\paragraph{Bootstrap Confidence.}
Confidence is computed in bootstrap mode and is used to control the acquisition regime rather than to terminate the run. After fitting a candidate mechanism, the domain-specific refinement backend is rerun on bootstrap resamples of the training split, and the resulting models are evaluated on a held-out validation split. Let $\hat{y}^{(b)}$ denote the prediction vector from bootstrap fit $b$. We compute bootstrap confidence from the mean coefficient of variation across validation predictions,
\[
\mathrm{conf}_{\mathrm{boot}}
=
1 - \frac{1}{N}\sum_{i=1}^{N}
\frac{\mathrm{std}_b\!\left(\hat{y}^{(b)}_i\right)}
{\left|\mathrm{mean}_b\!\left(\hat{y}^{(b)}_i\right)\right| + \varepsilon},
\]
clipped to $[0,1]$. In our setup, this confidence determines whether acquisition emphasizes hypothesis disambiguation or parameter refinement. When bootstrap confidence remains below the gating threshold, the system stays in a disagreement-driven regime and selects experiments using hypothesis-conditioned acquisition to separate competing mechanisms. Once confidence exceeds the threshold (0.9 in the paper configuration), the system switches to a refinement regime, where acquisition is driven by uncertainty sampling to reduce residual uncertainty within the current high-confidence mechanism class.

\paragraph{Memory and Final Selection.}
The memory injected into later prompts is a structured summary rather than a raw conversation transcript. It can include the full symbolic-regression history, the current best equation and its fit statistics, bootstrap confidence, mid-run ground-truth RMSLE when available, validated hypotheses that survived discriminating experiments, negative evidence for falsified forms, and a hypothesis scoreboard that tracks validated, failed, and uncertain structures. At budget exhaustion, \modelname{} performs a final symbolic fitting pass using the full symbolic-regression budget, reuses the final variable filter and hypothesis-family pool, and constructs a final candidate set from domain-specific optimization candidates, direct skeleton fits, and validated hypothesis survivors. The main model then arbitrates among these final candidates based on scientific plausibility and consistency with the observed data, rather than selecting solely on training $R^2$. If a mid-run candidate achieved a clearly better ground-truth validation score, that candidate is preserved as the final equation.

\subsection{Baselines}
\label{appendix_baselines}
All methods use the same oracle, task instance, difficulty, law version, and total experiment budget as \modelname{}. For the \textbf{LLM-only} and \textbf{Code-assisted LLM} conditions, we follow the same definitions and prompting/tool-use setups as in NewtonBench and refer readers to \cite{zheng2026newtonbench} for those implementation details. Below, we summarize the remaining comparison methods used in our benchmark suite.

\textbf{Random+PySR.} This method uses a non-adaptive experimental design followed by symbolic regression. It serves as the non-adaptive floor.

\textbf{BO+PySR.} This method uses Gaussian-process Bayesian optimization without LLM guidance. 

\textbf{BED+PySR.} This method performs Bayesian experimental design over a fixed hand-specified mechanism library. At each step, each candidate mechanism is fit to the current observations by nonlinear least squares in log-space, candidate experiments are sampled from the admissible bounds, and the next experiment is chosen to maximize the disagreement between the fitted mechanisms. After the budget is exhausted, the best-fitting library member is retained, and PySR is run as a final symbolic refinement stage. This uses the same general style of mechanistic library reasoning as the LLM pipeline, but with a fixed, predefined family set rather than dynamically generated hypotheses.

\textbf{GENIE3.} We evaluate GENIE3 on fixed \grn datasets collected under the same task budget and use the standard \texttt{GENIE3} implementation \citep{genie}, with a lightweight post-processing step to convert the output into the signed-graph format required by \grn evaluation.

\textbf{GIES.} We evaluate GIES on the same fixed \grn datasets using the standard \texttt{pcalg} implementation \citep{gies}, with intervention labels corresponding to the benchmark perturbation environments and the same signed-graph conversion.

\textbf{NOTEARS.} We evaluate the linear NOTEARS method on the same fixed \grn datasets using the reference \texttt{notears} implementation \citep{tears}, followed by the same signed-graph conversion used for benchmark evaluation.

\textbf{LLM-only and Code Assisted LLM.} We follow the implementation of these baselines as presented in NewtonBench~\cite{zheng2026newtonbench}. 

\subsection{Experimental Protocol}
\label{appendix_experimental_protocol}

Unless otherwise noted, all experiments use deterministic benchmark instances with zero oracle noise and matched task manifests. NewtonBench evaluates 12 domains across 3 difficulty levels, 3 law versions, and 3 seeds (324 runs total), with representative budget studies using 96 tasks. For \chem, representative manifests contain 36 tasks, with standard budgets \(B \in \{20,40,60,80,100\}\) and fixed-budget comparisons at \(B=60\). For \grn, representative manifests contain 36 tasks for budget studies and 18 for noise studies, using \(B \in \{10,20,50\}\) with fixed-budget comparisons at \(B=20\). NewtonBench fixed-budget comparisons also use \(B=20\), with extended studies reported up to \(B=100\). Evaluation is performed on held-out oracle outputs. Symbolic regression tasks report RMSLE-based recovery together with exact and symbolic recovery, while \grn reports edge \(F_1\), exact graph accuracy, and sign accuracy against the hidden graph. All LLM-based experiments use \texttt{GPT-4o-mini} as the primary reasoning model and \texttt{Qwen/Qwen2.5-7B-Instruct} for adaptive ensembling via local vLLM. The main model is used without task-specific fine-tuning, and ensemble sampling uses a temperature of 1.0 for structural diversity. Symbolic regression refinement uses PySR with 800 iterations plus direct fitting of candidate mechanism skeletons when available, while \grn uses signed-graph fitting with BFGS optimization. Bootstrap confidence from held-out validation fits is used only for acquisition-mode switching.

\subsection{Prompt Templates}
\label{app:prompts}
\begin{tcolorbox}[title=Hypothesis Generation (Symbolic Regression),colback=gray!5,colframe=black]
\tiny
\begin{verbatim}
GOAL: {goal}

DOMAIN: {domain}
PARAMETERS: {param_names}
{param_description}
OBJECTIVE TYPE: {objective_type}
OBJECTIVE DIRECTION: {objective_direction}
{optional_objective_profile}

FUNCTION SIGNATURE:
{function_signature}

DATA ({budget_pct}% budget used, {budget_remaining} experiments left):
{data_table}

Best symbolic equation so far:
{best_equation_fit}

{memory_str}
Current hypotheses: {current_hypothesis}
CURRENT PHASE: {current_phase}

TASK:
Generate one primary and 2-6 alternate EQUATION hypotheses that remain
plausible under current data.
Do NOT propose search regions in this step.
Focus only on plausible competing hypotheses and concise reasoning for
ambiguity.
\end{verbatim}
\end{tcolorbox}

\begin{tcolorbox}[title=Search-Region Proposal (Symbolic Regression),colback=gray!5,colframe=black]
\tiny
\begin{verbatim}
GOAL: {goal}

DOMAIN: {domain}
PARAMETERS: {param_names}
{param_description}
OBJECTIVE TYPE: {objective_type}
OBJECTIVE DIRECTION: {objective_direction}
{optional_objective_profile}

DISCOVERED LAW FUNCTION SIGNATURE (must match exactly):
{function_signature}

EXPERIMENTAL DATA ({budget_pct}% of budget used, {budget_remaining}
experiments remaining):
{data_table}

{best_equation_if_present}
{hypothesis_requirement_if_present}
Current best hypothesis: {current_hypothesis}
{memory_str}
{phase_instruction}
{discrimination_hints}

Based on this data, propose the most informative parameter regions to
explore next. Think carefully about what the data represents and what
experiments would most help discover the governing equation via disambiguating the competing hypotheses set.

search_regions must be a list of objects with this exact structure:
{"bounds": {"p1": [lo, hi], "p2": [lo, hi], ...},
 "n_experiments": 4,
 "priority": "high",
 "rationale": "why"}
\end{verbatim}
\end{tcolorbox}

\begin{tcolorbox}[title=Hypothesis-to-Executable Structure (Symbolic Regression),colback=gray!5,colframe=black]
\tiny
\begin{verbatim}
GOAL: {goal}

DOMAIN: {domain}
PARAMETERS: {params_str}

REQUIRED FUNCTION SIGNATURE (use EXACTLY this):
{function_signature}

EXPERIMENTAL DATA (raw values and [log10 values]):
{data_table}

CURRENT HYPOTHESIS: {current_hypothesis}
{previous_attempts_if_any}

Read the log10 columns to estimate exponents:
\Delta log10(measurement) / \Delta log10(param) \approx exponent.
Then propose a Python function 'discovered_law' that fits this
structural form.

RULES:
- Use free constants C0, C1, alpha, beta, ... — they will be fitted
  numerically.
- Use ONLY standard Python math (no imports). Powers: use ** not pow().
- Use the EXACT parameter names from the FUNCTION SIGNATURE above.
- Constants must appear in the return expression.
\end{verbatim}
\end{tcolorbox}

\begin{tcolorbox}[title=Hypothesis Generation and Region Proposal (Graph Recovery),colback=gray!5,colframe=black]
\tiny
\begin{verbatim}
GOAL:
{goal}

DOMAIN:
{oracle.param_description}

CURRENT BEST HYPOTHESIS:
{current_hypothesis}

RECENT DATA:
{store_to_table}

HYPOTHESIS FIT SUMMARY:
{lineage_summary}

MEMORY SUMMARY:
{memory_str}

CONFIDENCE SUMMARY:
{confidence_str}

ENSEMBLE GRAPH DISTRIBUTION:
{ensemble_summary}

{optional_diversity_requirement_block}

BUDGET REMAINING: {budget_remaining}
MAX EXPERIMENTS THIS ITERATION: {max_experiments_per_iter}

Return:
- 2-5 natural-language mechanism hypotheses with stable IDs like h1,
  h2, h3
- a primary_hypothesis_id naming the current best one
- 1-4 search regions for the next experiments
- confidence and done flag
\end{verbatim}
\end{tcolorbox}

\begin{tcolorbox}[title=Single-Hypothesis Sampling (Graph Recovery),colback=gray!5,colframe=black]
\tiny
\begin{verbatim}

GOAL:
{goal}

DOMAIN:
{oracle.param_description}

CURRENT BEST HYPOTHESIS:
{current_hypothesis}

RECENT DATA:
{store_to_table}

HYPOTHESIS FIT SUMMARY:
{lineage_summary}

MEMORY SUMMARY:
{memory_str}

CONFIDENCE SUMMARY:
{confidence_str}

BUDGET REMAINING: {budget_remaining}

Return exactly one plausible natural-language mechanism hypothesis,
with a short rationale and confidence.
Do not output graph edges or search regions.
\end{verbatim}
\end{tcolorbox}

\begin{tcolorbox}[title=Hypothesis-to-Structure Translation (Graph Recovery),colback=gray!5,colframe=black]
\tiny
\begin{verbatim}
Translate this single natural-language hypothesis into one signed graph.

hypothesis_id: {hypothesis_id}
text: {hypothesis_text}

Fill this schema exactly:
{
  "translation": {
    "hypothesis_id": "{hypothesis_id}",
    "rationale": "why this graph matches the hypothesis",
    "assumptions": ["optional assumption"],
    "edges": [
      {"src": "signal", "dst": "A", "sign":1},
      {"src": "A", "dst": "C", "sign":1}
    ]
  }
}

Requirements:
- dst must be one of {graphs_nodes}.
- The graph must contain a directed path from signal to {final_node}.
- If the hypothesis does not mention {final_node} explicitly, add the minimal
  faithful chain needed so the signal reaches {final_node}.
\end{verbatim}
\end{tcolorbox}

\section*{Acknowledgements}
This research was partially supported by the U.S. National Science Foundation (NSF) under Grant No. 2416728 and Autodesk Research. This work was supported by a Laboratory Directed Research \& Development (LDRD) project. 
This work was performed at the Center for Integrated Nanotechnologies, a U.S. Department of Energy Office of Science user facility. This article was authored by an employee of National Technology \& Engineering Solutions of Sandia, LLC under Contract No. DE-NA0003525 with the U.S. DOE. The employee retains all rights to the article and is solely responsible for its contents. The U.S. Government retains a non-exclusive, paid-up, irrevocable, worldwide license to publish or reproduce this work for government purposes. Public access will be provided in accordance with the DOE Public Access Plan: https://www.energy.gov/downloads/doe-public-access-plan.

\end{document}